\documentclass[sigconf,10pt,nonacm]{acmart}
\hypersetup{hidelinks}

\usepackage[linesnumbered,ruled,vlined]{algorithm2e}
\usepackage{array}
\usepackage{balance}
\usepackage{booktabs}
\usepackage{caption}
\usepackage{enumitem}
\usepackage{flushend}
\usepackage{float}
\usepackage{graphicx}
\usepackage{multirow}
\usepackage{mathtools}
\usepackage{pifont}
\usepackage{subcaption}
\usepackage{tabularx}
\usepackage{threeparttable}
\usepackage{xcolor}
\usepackage{tikz}
\lccode`/=`/

\definecolor{darkgreen}{RGB}{0,128,0} %
\definecolor{darkred}{RGB}{200,0,0}  %

\renewcommand\footnotetextcopyrightpermission[1]{} %

\newcommand{\name}{EdgeCraft}

\definecolor{nlpcolor}{RGB}{35,35,35}
\definecolor{cvcolor}{RGB}{25,84,140}
\definecolor{audiocolor}{RGB}{160,55,135}
\definecolor{sensingcolor}{RGB}{38,118,57}
\definecolor{tabularcolor}{RGB}{0,112,125}
\definecolor{mmcolor}{RGB}{132,31,45}

\newcommand{\bubble}[2]{%
\tikz[baseline=(char.base)]{
\node[
circle,
fill=#1,
draw=#1!65!black,
line width=0.2pt,
text=white,
font=\fontsize{3.7}{3.7}\selectfont\bfseries,
minimum size=8.5pt,
inner sep=0.1pt
] (char) {#2};
}
}

\newcommand{\nlptag}[1]{\bubble{nlpcolor}{#1}}
\newcommand{\cvtag}[1]{\bubble{cvcolor}{#1}}
\newcommand{\audiotag}[1]{\bubble{audiocolor}{#1}}
\newcommand{\sensetag}[1]{\bubble{sensingcolor}{#1}}
\newcommand{\tabtag}[1]{\bubble{tabularcolor}{#1}}
\newcommand{\mmtag}[1]{\bubble{mmcolor}{#1}}

\newcommand{\inlinebubble}[2]{%
\tikz[baseline=(char.base)]{
\node[
circle,
fill=#1,
draw=#1!65!black,
line width=0.2pt,
text=white,
font=\fontsize{4.2}{4.2}\selectfont\bfseries,
minimum size=9.5pt,
inner sep=0.1pt
] (char) {#2};
}
}

\newcommand{\inlinecvtag}[1]{\inlinebubble{cvcolor}{#1}}
\newcommand{\inlinenlptag}[1]{\inlinebubble{nlpcolor}{#1}}
\newcommand{\inlineaudiotag}[1]{\inlinebubble{audiocolor}{#1}}
\newcommand{\inlinesensetag}[1]{\inlinebubble{sensingcolor}{#1}}
\newcommand{\inlinetabtag}[1]{\inlinebubble{tabularcolor}{#1}}

\renewcommand{\arraystretch}{1.15}

\begin{document}

\title{\name: Automated Model Crafting for Edge IoT}

\author{Genglin Wang\textsuperscript{1},
Kaiwei Liu\textsuperscript{1},
Liekang Zeng\textsuperscript{1},
Wangsong Yin\textsuperscript{2},\\
Shangcheng Jin\textsuperscript{1},
Guoliang Xing\textsuperscript{1},
Zhenyu Yan\textsuperscript{1}
}

\affiliation{
\textsuperscript{1}Department of Information Engineering, The Chinese University of Hong Kong \city{Hong Kong SAR}
\country{China}\\
\textsuperscript{2}School of Computer Science, Peking University \city{Beijing} \country{China}\\
}

\renewcommand{\shortauthors}{Wang et al.}

\begin{abstract}
Machine learning (ML) increasingly powers Internet of Things (IoT) applications at the edge. Yet producing a deployable edge ML artifact for a specific scenario requires navigating a huge search space spanning data representation, model design, training on domain-specific data, and runtime customization. This workflow is fragmented and difficult to scale across diverse edge applications.

We present \name{}, an LLM-driven system that turns high-level intent into deployable edge ML artifacts. Building such a system raises two challenges: (1) How can an LLM be guided to find high-quality solutions that meet dynamic SLOs for task quality, latency, and energy? (2) How can trustworthy target-device verification be obtained at low cost? \name{} addresses these challenges with two designs. (1) A constraint-aware synthesis tree explores alternative candidates and uses measured SLO gaps to guide each improvement. (2) A multi-fidelity verifier progressively combines low-cost checks with full target-device verification to reduce verification cost while preserving reliable verification results. It also records verified failures for reuse, avoiding repeated device work. To support concurrency, \name{} provides a multi-tenant runtime that runs cloud training and target-device verification in parallel while isolating requests.
Across 50 public tasks, EdgeCraft exceeds the task-specific Reference in best-observed quality on 40 tasks and finds an SLO-feasible artifact on 45, with the two outcomes overlapping on 38 tasks. Moreover, \name{} achieves competitive performance on our self-collected SEN dataset, suggesting its generalizability to real-world IoT sensing tasks.

\end{abstract}

\maketitle
\hypersetup{pdfauthor={Genglin Wang, Kaiwei Liu, Liekang Zeng, Wangsong Yin, Shangcheng Jin, Guoliang Xing, Zhenyu Yan}}
\fancyhead[LE,RO]{}

\section{Introduction}\label{Introduction}

Machine learning (ML) increasingly powers edge Internet-of-Things (IoT) applications, such as agri-food~\cite{taneja2023artificial}, surveillance~\cite{srivastava2017safety}, wearable sensing~\cite{wearable_sensing}, and healthcare monitoring~\cite{qi2025alzheimer}.
However, producing a deployable model artifact for a specific IoT scenario still requires substantial expert effort. Although ML frameworks (e.g., PyTorch~\cite{pytorch_homepage_2026}) simplify training and inference, they do not automate the complete workflow. This difficulty arises from three factors.

\begin{figure}[!t]
\centering
\includegraphics[width=1.0\linewidth]{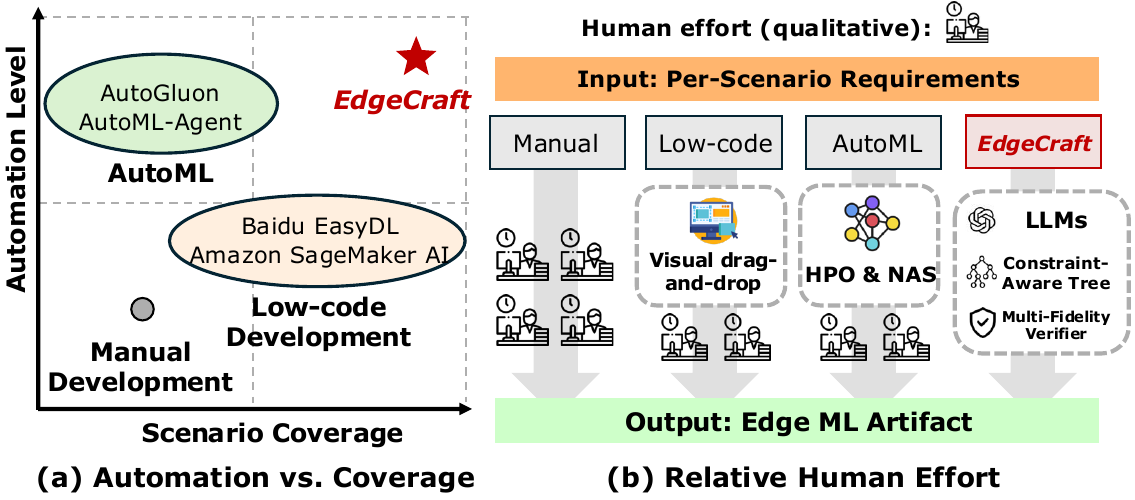}
\vspace{-17pt}
\caption{Qualitative comparison of \name{} with existing edge ML development paradigms in (a) automation and coverage, and (b) required human effort.}\label{fig:teaser}
\end{figure}

\begin{enumerate}[label=\textbf{(\alph*)}, wide=0pt, leftmargin=0pt, nosep]
\item \textbf{Diverse applications and service-level objectives.}
Edge ML spans vision, language, audio, sensing, tabular, and multimodal applications. Their service-level objectives (SLOs) differ substantially~\cite{yin2025elasticondevicellmservice}, including task quality (e.g., accuracy and mAP) and system performance (i.e., latency and energy). Table~\ref{tab:edge_ml_slo} presents representative examples: visual inspection prioritizes recall or AUROC, wearable sensing must limit energy consumption, and always-on keyword spotting requires low latency.

\item \textbf{A huge search space.}
Developers must make coupled choices about data representation, model family, training recipe, and inference runtime. Even within edge vision, model families such as YOLO~\cite{redmon2016you,ultralytics2026} and MobileNet~\cite{MobileNet,MobileNetV2,MobileNetV3} expose substantially different accuracy-latency-energy trade-offs. General model libraries further enlarge the space of available architectures~\cite{torchvision2016,rw2019timm}. The design space therefore extends far beyond a fixed model zoo.

\item \textbf{Fragmented execution environments.}
Edge devices span hardware families such as Raspberry Pi~\cite{raspberrypi} and NVIDIA Jetson~\cite{nvidia_jetson_2026}, with different compute capabilities and software stacks. Their deployment stacks combine different inference runtimes, such as PyTorch~\cite{pytorch_homepage_2026}, ONNX~\cite{onnx_github}, and TensorRT~\cite{tensorrt_homepage}. A model that trains successfully may still fail during compilation on the target device. Latency and energy also depend on the devices and runtimes~\cite{zhang2021nnmeter}.

\end{enumerate}

\begin{figure*}[!htbp]
\centering
\includegraphics[width=1.0\linewidth]{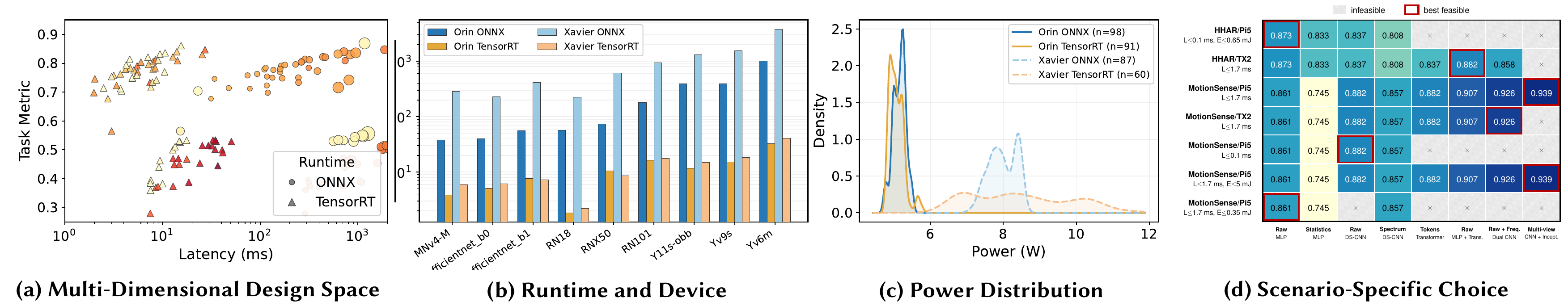}
\vspace{-17pt}
\caption{(a) Task quality, latency, and power vary across model–runtime configurations. Marker shape and size denote runtime and peak memory, while color denotes measured power from yellow (low) to red (high). (b–c) Measured p95 latency and power vary substantially across device–runtime combinations. (d) The HAR design changes with the dataset, device, and SLOs; hatching marks SLO violations and red boxes mark winners.}
\label{fig:obs1}
\vspace{-5pt}
\end{figure*}

These factors are tightly coupled. Changing any of these factors can change the best feasible design for a specific IoT scenario. For example, a change in data representation can alter accuracy and exported operators; changing the runtime can affect latency and energy. Moreover, satisfying one SLO may degrade another. Our measurements in Figure~\ref{fig:obs1}(d) also confirm this. Therefore, developers must navigate a scenario-dependent design space and repeatedly design, measure, and revise. Existing automated approaches also provide limited automation and scale poorly across scenarios (Figure~\ref{fig:teaser}).

\noindent\textbf{Model Crafting as a Service (MCaaS).}
These limitations motivate a new service abstraction for edge ML development, which we call \textit{Model Crafting as a Service (MCaaS)}\footnote{Model crafting refers to designing, training, exporting, and delivering a deployable edge ML artifact on the edge, i.e., a model-centric process.}. Given a user intent, dataset, target device, and service-level objectives, MCaaS seeks to return a deployable ML artifact under a limited search budget (e.g., time or trials). MCaaS is verification-grounded, which means that search decisions rely on results from compiling, running, and measuring candidates on target devices. If no SLO-feasible artifact is found, it reports the best verified candidate together with its remaining constraint gaps. MCaaS therefore defines an end-to-end service contract between high-level application requirements and physically validated edge ML artifacts. As a service, MCaaS pools scarce cloud and edge resources across isolated tenant requests and reuses verified compatibility rules across jobs.
Recent LLM agents provide a promising foundation for this vision by interpreting open-ended intent and coordinating multi-stage development planning and
workflows~\cite{karpathy_autoresearch_2026,novikov2025alphaevolve,jiang2025aide,hong2025datainterpreter, IoT-Brain,romera2024funsearch,yang2025proagentharnessingondemandsensory,panchal2026mosaicruntimeefficientmultiagentembodied,zhang2025open3dvqabenchmarkcomprehensivespatial}. However, their general knowledge provides useful design priors rather than reliable knowledge of physical behavior. Strong generative capabilities therefore do not by themselves yield a reliable ML synthesis process.

\begin{table}[t]
\centering
\footnotesize
\setlength{\tabcolsep}{8pt}
\renewcommand{\arraystretch}{1.08}
\resizebox{\linewidth}{!}{
\begin{tabular}{c|c}
\toprule
\textbf{Edge/Mobile ML App.} & \textbf{Service-Level Optimization Objective} \\
\midrule
Camera surveillance~\cite{liu2018ano_pred}
& Max. AUC, acceptable latency  \\

Visual inspection~\cite{MVTecAD1,MVTecAD2}
& Max. recall/AUROC, acceptable latency \\

Drone-based inspection~\cite{VisDrone}
& Max. mAP, low latency, bounded energy \\

Wearable HAR~\cite{he2026egologegocentricfinegraineddaily}
& Max. accuracy, low latency, bounded energy \\

Always-on keyword spotting~\cite{speechcommandsv2}
& Max. accuracy, very-low latency \\

AIOps anomaly detection~\cite{telelogs}
& Max. F1/AUROC, very-low latency  \\

\bottomrule
\end{tabular}
}
\vspace{5pt}
\caption{Representative service-level optimization objectives for edge ML applications, reflecting different priorities among task quality, latency, and energy.}
\label{tab:edge_ml_slo}
\vspace{-8pt}
\end{table}

\noindent\textbf{Challenges.}
Specifically, realizing MCaaS raises two core design challenges. First, \textit{how can the system steer ML solution candidates toward application-specific SLOs in a huge search space?} Generating valid edge ML code is comparatively straightforward, but aligning the resulting artifact with application-specific SLOs remains challenging.
Our preliminary study (\S~\ref{subsec:preliminary}) confirms that the best solution varies across data distributions, target devices, and SLO requirements. Because an optimal solution may be scenario-specific, MCaaS must provide SLO-guided development and return the highest-quality artifact satisfying the user's requirements. Second, \textit{how can the system obtain trustworthy target-device verification results without fully evaluating every candidate?} During this search, a revision (e.g., adding a module or changing the data representation) may improve task quality but violate a latency or energy SLO, and its effect may become clear only after target-device verification. Prior systems therefore rely on a ``verify-before-commit'' methodology or device-specific calibration to obtain trustworthy physical feedback or calibrated predictions~\cite{IoT-Brain,zhang2021nnmeter,liu2026instmeter}. Fully training, deploying, and measuring every candidate is expensive, while cheaper checks provide verification results with different levels of reliability.
Beyond these two design challenges, operating MCaaS also requires managing shared cloud GPUs, edge devices, and verified compatibility rules. Each request has its own private, evolving search tree, while its training and verification jobs share a pool of GPUs and edge devices. The service coordinates these jobs and stores rules derived from reproduced device–runtime failures so that later requests can reuse them.

\noindent\textbf{\name{}.}
In this paper, we propose \name{}, an MCaaS system that achieves fully automated edge ML synthesis (Figure~\ref{fig:teaser}).
Its central principle is a separation between proposal and decision authority: LLMs propose and prioritize candidates, while verification determines which candidates may be accepted or rejected. \name{} has two key designs:
(a) \textit{Constraint-aware synthesis tree.} \name{} organizes edge ML development as a constraint-aware synthesis tree, where each node stores a candidate and its verification results. \name{} records each node’s parent–child history and expresses the measured task quality, latency, and energy as gaps or slack relative to the requested SLOs. The tree uses these results to generate and refine child candidates toward the remaining objectives.
(b) \textit{Multi-fidelity verifier}. To produce these verification results efficiently, the Multi-Fidelity Verifier progressively evaluates each candidate through static checks, low-cost device measurements, and full training with target-device verification. Decision authority depends on how each verification result is obtained: proxy results guide exploration, while verified static incompatibilities or calibrated physical measurements may prune a candidate. Only full evaluation can establish an artifact as feasible, and inconclusive low-cost verification results advance to full verification. To operate both mechanisms across concurrent requests, \name{} provides \textit{Cross-Tenant Shared Services}. These services coordinate training and execution jobs over pooled cloud GPUs and edge devices and store verified compatibility rules for later reuse.

\noindent\textbf{Evaluation.}
We construct a comprehensive benchmark spanning diverse modalities, devices, and runtimes (Table~\ref{tab:mcaas-benchmarks}).
Across a comprehensive 50-task public benchmark, EdgeCraft exceeds the task-specific Reference in best-observed quality on 40 tasks and finds an SLO-feasible artifact on 45, with the two outcomes overlapping on 38 tasks.
Compared with agentic approaches, \name{} produces SLO-feasible artifacts for 2.4$\times$ as many tasks. Its Multi-Fidelity Verifier reduces verification time by 25.7\% while preserving the selected artifact and its task quality. On the private SEN dataset, \name{} shows competitive performance, demonstrating its generalizability to a real-world wearable IoT sensing task.

\noindent \textbf{Contributions.} We summarize our contributions below:

\begin{enumerate}[label=\(\bullet\), wide=0pt, leftmargin=*, nosep]
\item We introduce the first MCaaS system, coupling open-ended code synthesis with target-device verification to produce verified artifacts for user-specified tasks, devices, and SLOs.

\item Our constraint-aware tree preserves candidates and uses verified quality and SLO gaps to revise branches; multi-fidelity probes conditionally reject, while only full verification accepts artifacts.

\item A comprehensive benchmark covering 50 public datasets across six modalities and a self-collected sensor task confirms \name{}'s strong potential and generalizability across edge/mobile ML crafting tasks.

Code is available.\footnote{\url{https://github.com/genglinWang/EdgeCraft}.}

\end{enumerate}

\vspace{-8pt}
\section{Background and Motivation}
\subsection{Edge ML Development Characteristics}

\noindent\textbf{Edge ML is pervasive.}
Edge ML has become a pervasive building block in IoT applications. These applications execute ML locally to protect sensitive data and remain functional under limited connectivity~\cite{singh2023edge,zhou2019edge}. Crucially, although edge ML is pervasive, each edge ML solution remains inseparable from its specific scenario. There is no universally best solution: its suitability depends jointly on the application task, target device, and service-level objectives. Developers must adapt available model families, training recipes, and inference runtimes to build a deployable solution that satisfies scenario-specific SLOs.

\noindent\textbf{Why do we need end-to-end automation?}
Manual development has not scaled with the growth and diversity of edge AI applications. Developers must repeatedly design, train, export, deploy, and measure a candidate before deciding how to revise it. Existing automation remains fragmented. For instance, AutoML frameworks automate model selection and training within predefined search spaces~\cite{he2021automl, AutoGluon_agtabular,h2o_python_2016,10.1145/3450268.3453520}, but typically stop before target-device deployment and measurement. LLM-based program synthesis can generate training programs~\cite{novikov2025alphaevolve,jiang2025aide,shen2025autoiot,yang2024embedgenius}, yet does not reliably close the loop through target-device verification. End-to-end automation is therefore needed to connect these stages, return verification results to subsequent revisions, and continuously refine each candidate for its deployment scenario.

\subsection{Preliminary Study}\label{subsec:preliminary}
We investigate three questions: how strongly edge ML solutions depend on their scenarios, whether LLM-generated ML programs survive target-device execution, and whether LLMs provide a useful prior for navigating the solution space.

\noindent\textbf{Scenario-dependent solutions and validation cost.}
We obtain 336 device--runtime profiles for 107 vision models from timm~\cite{rw2019timm}, torchvision~\cite{torchvision2016}, and Ultralytics~\cite{ultralytics2026} on Jetson AGX Orin and Jetson Xavier NX using ONNX Runtime and TensorRT. Figure~\ref{fig:obs1}(a) shows that quality, latency, and power do not improve together across configurations; increasing model scale therefore does not reliably identify the best deployable solution. The deployment configuration alone can dominate performance: for ResNet-18, p95 latency differs by 122$\times$ across the four device--runtime combinations, while the highest measured power is 2.4$\times$ the lowest (Figures~\ref{fig:obs1}(b) and~\ref{fig:obs1}(c)).
Figure~\ref{fig:obs1}(d) shows the same scenario dependence on HHAR~\cite{hhar} and MotionSense~\cite{motionsense}. Across seven dataset--device--SLO scenarios drawn from the same candidate pool, five different designs achieve the highest SLO-feasible quality, and no design wins more than twice. These winners combine raw, frequency, and multi-view representations with MLP, convolutional, Transformer, and Inception components. Thus, neither model scale nor task quality alone determines the appropriate edge ML solution.

Exhaustively identifying these winners is costly. Exploring these
alternatives can consume substantial GPU time, with prior architecture searches
reporting thousands of GPU-days~\cite{ASAP}; surviving artifacts must then be
verified on their target devices. Edge ML synthesis must therefore explore
selectively, reserving full training and target-device verification for
promising candidates.

\noindent\textbf{Observation \#1:}
\textit{Edge ML solutions are scenario-dependent and costly to validate. Synthesis therefore requires an efficient agent to navigate the open-ended solution space.}

\begin{figure}[!t]
\centering
\begin{subfigure}[t]{0.48\linewidth}
\centering
\includegraphics[width=\linewidth]{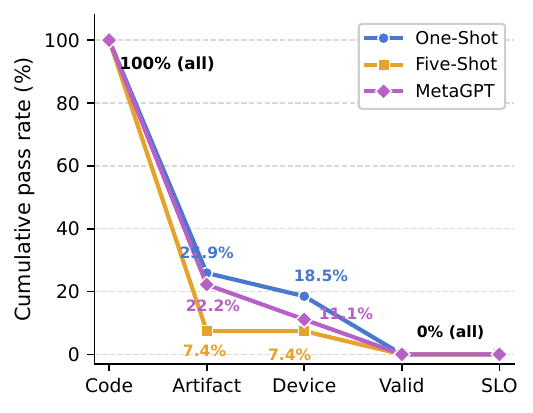}
\caption{Pass funnel from valid code to SLO satisfaction.}
\label{fig:obs2_funnel}
\end{subfigure}
\hfill
\begin{subfigure}[t]{0.48\linewidth}
\centering
\includegraphics[width=\linewidth]{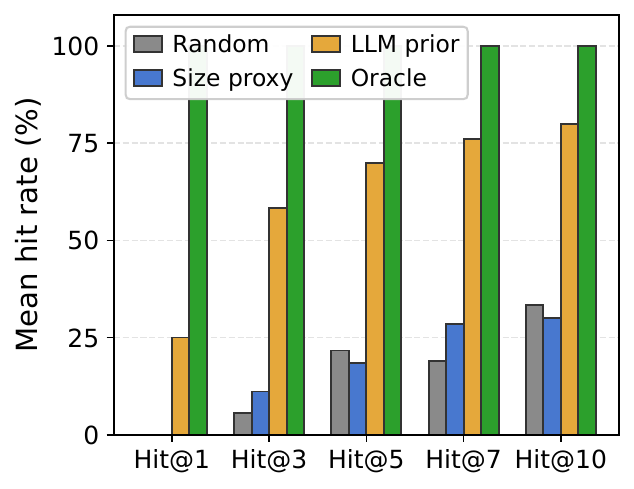}
\caption{Sample efficiency of candidate selection.}
\label{fig:obs3_hitk}
\end{subfigure}
\vspace{-3pt}
\caption{Motivation studies of LLM-driven synthesis.
(a) LLM-driven synthesis fails before functional validation on the target device. (b) LLM-guided search identifies promising candidates more
efficiently.}
\label{fig:motivation_study}
\end{figure}

\noindent\textbf{LLM agents do not by themselves synthesize a reliable edge ML model.}
We examine whether existing LLMs and LLM agents can produce deployable edge ML
implementations using all 27 benchmark tasks mapped to Jetson Xavier NX.
We evaluate one-shot and five-shot GPT-5.4~\cite{openai_gpt5_chat} and
MetaGPT~\cite{hong2024metagpt}, yielding 81 candidates. Each
candidate is executed unchanged in an isolated workspace on the physical
device.
We record five
cumulative stages: valid code, requested-runtime artifact production,
target-device execution, evaluator-validated output, and satisfaction of all
active SLOs.
Figure~\ref{fig:obs2_funnel} shows that all 81 candidates produce valid code,
but only 15 (18.5\%) produce the requested artifact, and 10 (12.3\%) execute
on the target device. None produces evaluator-validated output or an
SLO-feasible candidate. Although two candidates report measurements within
their numerical SLO thresholds, their outputs fail functional validation and
are therefore excluded. Failures commonly result from unavailable device
libraries or runtime interfaces and incompatible export or precision
assumptions. Several failures recur across independently generated candidates,
repeatedly consuming target-device verification time.

\noindent\textbf{Observation \#2:}
\textit{LLM-generated candidates rarely produce valid target-device outputs.
Synthesis must therefore apply target-device verification and avoid repeating known failures.}

\noindent\textbf{LLMs provide useful quality priors when provided with target-device verification.}
We test whether an LLM can prioritize promising candidates without determining
their physical feasibility. We reuse the profiled models from
Observation~\#1, for which task quality and target-device performance have
both been measured. Under a 20-trial budget, we compare Random selection, a Size-Proxy heuristic, an LLM Prior, and an Oracle. Size-Proxy orders candidates from smaller to larger using standard model-family size, as a lightweight proxy for computational cost. The LLM ranks
candidates only by expected task quality, while measured device latency
determines feasibility. Hit@$K$ measures how often a strategy finds the
highest-quality feasible candidate within its first \(K\) trials.
Figure~\ref{fig:obs3_hitk} shows that the LLM Prior finds strong candidates
earlier. It achieves
80\% Hit@10, compared with 33\% for Random and 30\% for Size-Proxy; the
Oracle achieves 100\%. Its rankings also correlate positively with measured
task quality, with Spearman coefficients from 0.695 to 0.865. Thus, the LLM
guides where to search, while target-device measurements determine which
candidates are feasible.

\noindent\textbf{Observation \#3:}
\textit{An LLM quality prior improves search efficiency by directing limited
trials toward promising candidates, while target-device measurements determine feasibility.}

\subsection{Problem Formulation}

To address fragmented and device-specific edge ML development, we formulate
Model Crafting as a Service (MCaaS): a mapping from user intent, dataset,
target device, and SLOs to a deployable edge ML artifact within a bounded
development budget. Existing paradigms automate parts of this process but
differ in synthesis scope, target-device verification, and cross-request
reuse.

\noindent\textbf{Service model.}
MCaaS is a multi-tenant cloud service backed by a training cluster and a managed pool of physical edge devices. The GPU cluster supports scalable candidate training across concurrent requests, while the provider maintains an extensible catalog of edge device families, such as NVIDIA Jetson, making provider-side target-device verification practical. The tenant specifies the target device, while the service selects a qualified runtime and returns an artifact with its measured target-device performance.

\noindent\textbf{Request and objective.}
Each tenant request is represented as \(R_i=(u_i,D_i,h_i,C_i,B_i)\), where
\(u_i\) is the user intent, \(D_i\) is the dataset, \(h_i\) is the target
device, \(C_i\) specifies the task-quality objective and applicable physical
SLOs, and \(B_i\) limits the number of evaluated candidates or wall-clock
time. The service seeks the highest-quality valid candidate satisfying these
SLOs:
\begin{equation*}\footnotesize
x_i^\star=\arg\max_{x\in\mathcal{X}_i(B_i)}q_i(x)
\;\mathrm{s.t.}\;\operatorname{Valid}_{h_i}(x)=1,
\;x\text{ satisfies SLOs in }C_i .
\end{equation*}
Here, \(\mathcal{X}_i(B_i)\) contains the candidates explored within the
budget, and \(q_i(x)\) is the request-specific quality score, oriented so
higher is better. \(\operatorname{Valid}_{h_i}(x)\) requires a qualified-runtime artifact to be generated, executed on \(h_i\), and functionally validated. Physical SLOs include latency and energy. The service returns the selected artifact and its
measurements; if no candidate satisfies every SLO, it returns the best candidate and its remaining constraint gaps. This objective follows
hardware-aware model development, which maximizes predictive quality under
device constraints~\cite{cai2020onceforalltrainnetworkspecialize,
yang2018netadaptplatformawareneuralnetwork,
tan2019mnasnetplatformawareneuralarchitecture}.

\noindent\textbf{Workflow principles.}
Each candidate proceeds through development, cloud training and export, and
target-device verification. Although edge platforms can support
training~\cite{PieBridge}, repeated training is slow and occupies devices
needed for target-device verification. These stages are sequential for one candidate,
but concurrent requests enable two forms of sharing. \emph{Pipeline sharing}
overlaps cloud work with target-device verification across requests, while
\emph{rule sharing} lets sanitized, verified compatibility
rules benefit future requests. Tenant datasets, code, models, artifacts,
and development states remain isolated.

\section{\name{} Design}\label{sec:design}

\subsection{Overview}

\noindent\textbf{Design principle.}
\name{} separates proposal from decision authority: the LLM explores the open
code space and chooses what to try next, while the verifier determines which
candidates may be rejected or accepted. The verifier balances reliability and
cost by combining low-cost checks with higher-cost target-device execution.
Each result is retained and shapes later proposals and decisions.

\begin{figure}[!t]
\centering
\includegraphics[width=0.96\linewidth]{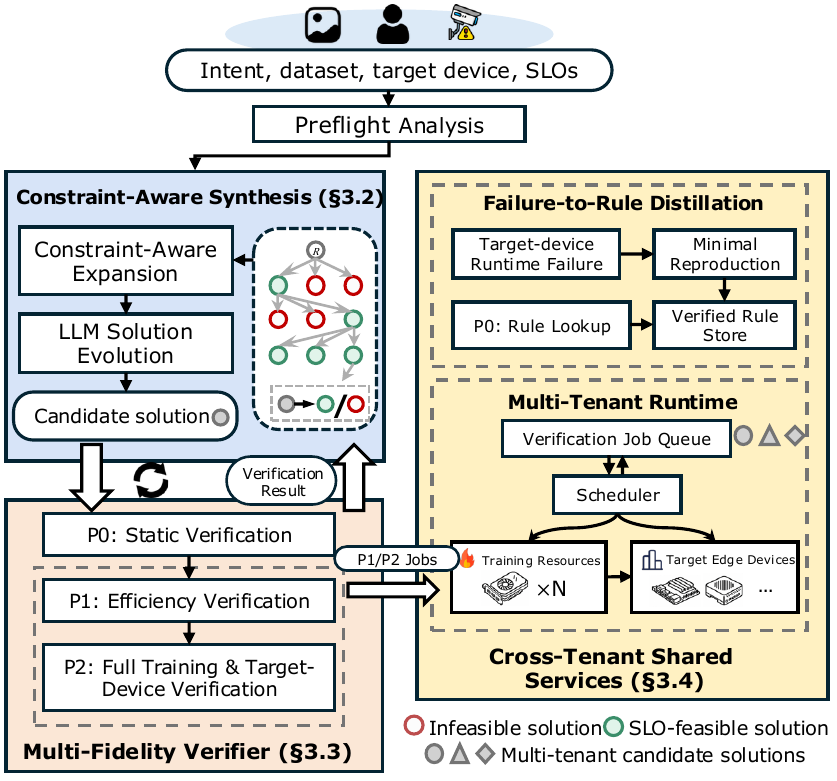}
\vspace{-7pt}
\caption{Overview of \name{}. }
\label{fig:system_overview}
\end{figure}

\noindent\textbf{Architecture.}
Figure~\ref{fig:system_overview} shows \name{}'s two core designs and the shared services.
(1) \textit{Constraint-Aware Synthesis Tree} (\S\ref{sec:synthesis_tree}). For each specific task, the tree organizes alternative candidates and evolves them toward the ML artifact that satisfies the user's requirements. (2)~\textit{Multi-Fidelity Verifier} (\S\ref{sec:multi_fidelity_verifier}) progressively evaluates each candidate through a low-cost P0 static check, a low-cost P1 target-device measurement, and a high-cost P2 full training and target-device execution. The results include the physical information required for decisions, such as latency, energy, and runtime status. They determine whether the candidate is eligible for further exploration and are returned to the LLM to guide expansion and revision. (3)~\textit{Cross-Tenant Shared Services} (\S\ref{sec:shared_services}). A shared runtime schedules overlapping training and target-device execution pipelines for all tenants across pooled cloud GPUs and target edge devices. \name{} also conducts Failure-to-Rule Distillation, which converts reproduced device--runtime failures into reusable knowledge that helps avoid repeated failures across tenants.

\noindent\textbf{Request lifecycle.} Each MCaaS request creates a private synthesis tree. The
LLM reads earlier verification results, selects a live branch, and proposes the
next candidate. The candidate then proceeds through P0 static checks, P1 target-device measurements, and P2 full training and execution. Reliable P0 or P1 results can prune a candidate, while only P2 can accept a candidate. Every result is written back to the tree and guides the
next expansion. Cross-Tenant Shared Services schedule P1 and P2 jobs and reuse verified compatibility rules.
Algorithm~\ref{alg:constraint-aware-synthesis} summarizes this lifecycle.

{%
\SetAlgoSkip{}
\setlength{\interspacetitleruled}{1pt}
\SetInd{0.35em}{0.70em}
\SetNlSkip{0.25em}
\newcommand{\CompactAlgoComment}[1]{{\scriptsize\ttfamily #1}}
\SetCommentSty{CompactAlgoComment}
\begin{algorithm}[!t]
\footnotesize
\KwIn{Request $R=(\text{task},\text{data},\text{device},\text{SLOs})$; budget $B$}
\KwOut{Best fully verified SLO-feasible artifact}
$\mathcal{T} \leftarrow \textsc{InitializeTree}(R,\textsc{Preflight}(R))$\;
\While{$\textsc{CandidateCount}(\mathcal{T}) < B$}{
$\mathcal{P} \leftarrow \textsc{LiveBranches}(\mathcal{T})$\tcp*[r]{Verified and not pruned.}
\If{$\mathcal{P}=\varnothing$}{\textbf{break}\;}
$S \leftarrow \textsc{SummarizeResults}(\mathcal{T})$\;
\tcp{Per branch: quality progress, signed normalized SLO gaps $\mathbf{z}$ ($+$: unmet; $-$: slack), and inherited components.}
$p \leftarrow \textsc{SelectBranch}(\mathcal{P},S)$\tcp*[r]{Verified quality progress and signed SLO gaps $\mathbf{z}$.}
\Repeat{$v$ is executable, extends $p$, and references only verification results in $\mathcal{T}$}{
$v \leftarrow \textsc{ExpandBranch}(p,S)$\tcp*[r]{Bounded reject/retry.}
}
$(r,\ell) \leftarrow \textsc{VerifyNode}(v)$\;
\tcp{Progress through P0–P2 until $\textsc{CanReject}(r,\ell)$ holds or full evaluation (\S\ref{sec:multi_fidelity_verifier}).}
$\textsc{RecordResult}(\mathcal{T},v,r)$\tcp*[r]{Persist for later expansions.}
\eIf{$\textsc{CanReject}(r,\ell)$}{
$\textsc{PruneBranch}(\mathcal{T},v)$\;
}{
\If{$\ell\ \text{is full evaluation}\ \land\ \max_j z_j(v)\le 0$}{
$\textsc{AcceptBranch}(\mathcal{T},v)$\;
}
}
}
\Return{\textsc{SelectOutput}$(\mathcal{T})$, $\mathbf{z}_{\mathrm{out}}$}\;
\caption{Constraint-aware synthesis tree.}
\label{alg:constraint-aware-synthesis}
\end{algorithm}
}

\subsection{Constraint-Aware Synthesis Tree}
\label{sec:synthesis_tree}

\noindent\textbf{Tree state and initialization.}
For request $i$, the private tree is $\mathcal{T}_i=(\mathcal{V}_i,\mathcal{E}_i)$, where $\mathcal{V}_i$ is the set of nodes and $\mathcal{E}_i$ records their parent–child expansion relations. Upon receiving the request, \name{} initializes the tree as follows:

\noindent $\bullet$ \textsc{Preflight}$(R)$
summarizes the dataset schema, target-device capabilities, and qualified
runtimes.

\noindent $\bullet$ \textsc{InitializeTree} generates and verifies the initial candidates. Each node $v$ stores a candidate $x_v$ and a
verification state.

The verification state records the candidate's verification results, including its inherited components, quality progress, and signed SLO gaps. Each edge $(u,v)$ records the change from parent to child. $\mathcal{P}$ denotes the set of live branches. A verified, unpruned node is live. An accepted node may remain live when its slack leaves room for a higher-quality child candidate.
To express different constraints in one verification summary, \name{} converts each constrained metric into a signed gap relative to its requested SLO. For metric $m_j$ with target $\tau_j$ and scale $s_j=\max(|\tau_j|,\epsilon)$, the gap is
\vspace{-3pt}
\begin{equation*}\footnotesize
z_j =
\begin{cases}
(\hat m_j-\tau_j)/s_j, & m_j\leq\tau_j,\\
(\tau_j-\hat m_j)/s_j, & m_j\geq\tau_j.
\end{cases}
\end{equation*}

\begin{figure}[!t]
\centering
\includegraphics[width=0.98\linewidth]{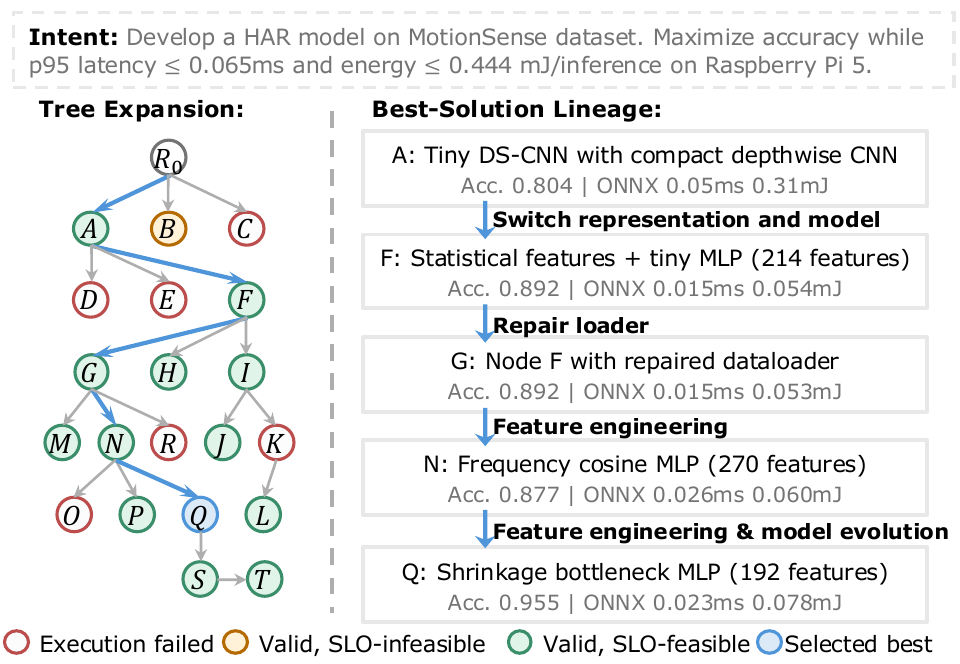}
\vspace{-6pt}
\caption{An illustrative synthesis tree on MotionSense. The left
panel shows the tree structure and its verification results; the
right traces successive mutations.}
\label{fig:tree}
\end{figure}

A positive $z_j$ means that the constraint remains unmet. A negative value is slack. Task-quality progress is tracked separately. Together, these values show what a branch has improved and which objective should guide its next revision. For example, a
child candidate may improve accuracy but retain a positive latency gap. Once that gap
becomes negative, the branch has slack for a more accurate model.
Figure~\ref{fig:tree} shows these states on competing MotionSense branches.

\noindent\textbf{Verification-guided expansion.}
At each iteration, \name{} expands the synthesis tree as follows:

\noindent$\bullet$ \textsc{SummarizeResults}$(\mathcal{T})$ builds a summary $S$ for the branches in $\mathcal{P}$. It contains quality progress and signed gaps $\mathbf{z}$.

\noindent$\bullet$ \textsc{SelectBranch} asks
the LLM to choose a parent $p$ from $\mathcal{P}$ and cite the verification results supporting that choice.

\noindent$\bullet$ \textsc{ExpandBranch} asks the LLM to produce one child candidate. \name{} admits the candidate only if it is executable, extends
$p$, and cites verification results stored in $\mathcal{T}$.

An invalid proposal receives a violation note and is retried up to a fixed bound. If the bound is reached, the attempt ends and control returns to the live branch set. Only an admitted candidate consumes one unit of $B$.

\noindent\textbf{Tree update and completion.}
\textsc{VerifyNode}$(v)$ (\S~\ref{sec:multi_fidelity_verifier}) returns a verification result $r$ and its fidelity $\ell$.
\name{} records this result before selecting the next branch.
\textsc{CanReject}$(r,\ell)$ is true only when the returned verification result has
rejection authority; Section~\ref{sec:multi_fidelity_verifier} defines that
authority. A rejected branch is pruned, while only a full evaluation with
$\max_j z_j(v)\leq 0$ registers an accepted artifact. Every recorded verification result,
including results from pruned branches, remains in the tree and shapes the next verification summary
$S$. \name{} returns the highest-quality accepted artifact. If none has been accepted, MCaaS returns the best verified candidate and its remaining gaps.

\vspace{-5pt}
\subsection{Multi-Fidelity Verifier}
\label{sec:multi_fidelity_verifier}

\noindent\textbf{Check, measure, confirm.}
Full training and target-device verification provide comprehensive information about task quality, latency, and energy, but they require a large number of GPU hours. To obtain such information without target-device execution, recent work has shown that latency can be predicted for specific devices, runtimes, and model types~\cite{zhang2021nnmeter,liu2026instmeter}. However, open-ended generation can produce candidates beyond that coverage. Our preliminary study
(\S~\ref{subsec:preliminary}) shows that many candidates can be rejected before full evaluation. \name{} therefore follows \textit{Check $\rightarrow$ Measure
$\rightarrow$ Confirm} through three probes, P0–P2. This section defines \textsc{VerifyNode}$(v)$ and
\textsc{CanReject}$(r,\ell)$ in Algorithm~\ref{alg:constraint-aware-synthesis}.

\begin{figure}[!t]
\centering
\includegraphics[width=0.98\linewidth]{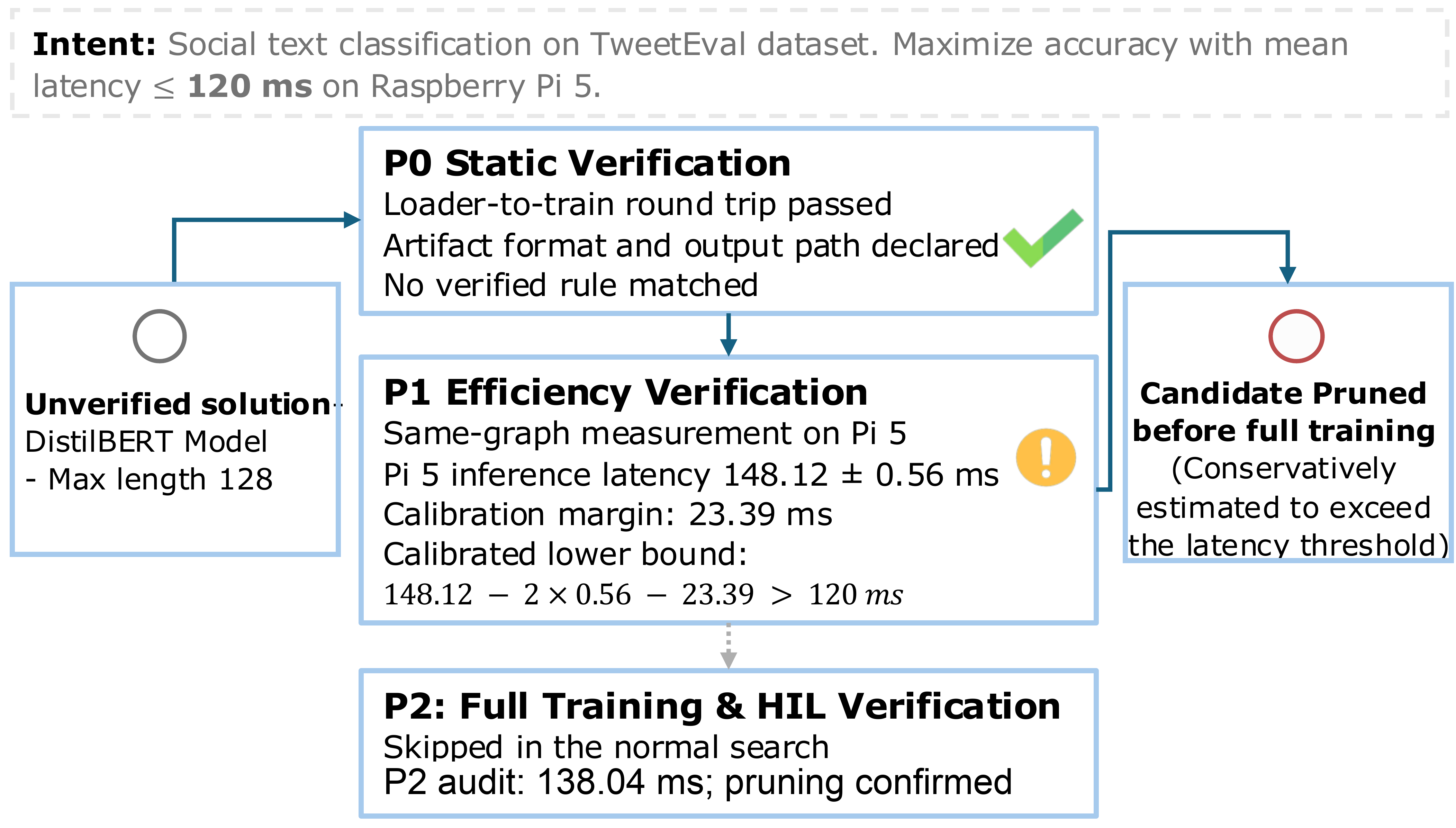}
\vspace{-6pt}
\caption{Calibrated device measurements prune a candidate before full training.}
\label{fig:multi-fidelity-trace}
\end{figure}

\noindent\textbf{\textsc{VerifyNode}$(v)$.} It evaluates a candidate using three probes.

\noindent $\bullet$ P0 checks component and artifact contracts and queries verified compatibility rules (\S~\ref{sec:shared_services}). This static probe requires neither training nor device execution.

\noindent $\bullet$ P1 constructs the candidate deployment
graph before full training and directly measures latency and, where applicable,
energy on the target device.

\noindent $\bullet$ P2 fully trains the candidate, exports its artifact, and
verifies both task quality and physical performance.

P0 and P1 may reject only under the conditions below; only P2 can accept an artifact. Every result, including the constraint gap of a rejected candidate, returns to the tree.

\noindent\textbf{P1 and calibration.}
P1 gates the training cost of graph-preserving refinements: it measures an initialized artifact before full training, whereas P2 measures the final trained artifact. A calibration context consists of the deployment-graph hash, device–runtime environment, physical metric, and execution setting. Calibration is reused only when this complete context matches; within it, calibration captures empirical P1--P2 and run-to-run variation. Graph-changing revisions, including quantization and export rewriting, create a new context and proceed to P2. The first candidate in a context completes both probes to create a P1–P2 pair. Later candidates or requests with a matching context use the largest completed difference as $\epsilon_{\mathrm{cal}}$. Only earlier pairs inform a decision; each new pair is added afterward.

\noindent\textbf{\textsc{CanReject}$(r,\ell)$.}
At P0, context-stable artifact and load incompatibilities, such as an unsupported ONNX IR version, may reject directly. Compilation- or execution-stage rules require the same deployment-graph hash and compiler context. Incomplete matches provide guidance only. At P1, an earlier calibration pair must match the complete calibration context defined above. For upper-bounded physical metrics—i.e., latency and energy—P1 rejects only
when
\begin{equation*}\footnotesize
\hat m_j-2\sigma_j-\epsilon_{\mathrm{cal}}>\tau_j.
\end{equation*}
The left-hand side is a conservative lower bound: it subtracts run-to-run
variation $2\sigma_j$ and the largest completed P1–P2 difference
$\epsilon_{\mathrm{cal}}$ from the P1 mean $\hat m_j$. P1 rejects only if this
bound still exceeds the SLO $\tau_j$. Data-dependent training proceeds to P2 because task-quality metrics, such as accuracy and AUROC, cannot be reliably predicted for a new dataset whose modality and data distribution may differ.

\noindent\textbf{Example.}
Figure~\ref{fig:multi-fidelity-trace} shows the decision for TweetEval. Earlier
matching pairs give $\epsilon_{\mathrm{cal}}=23.39$ ms. DistilBERT’s P1 latency
is $148.12\pm0.56$ ms on Pi~5, producing a conservative lower bound of
$123.61$ ms. Because this remains above the $120$ ms SLO, P1 rejects the
candidate. Continuing the candidate through P2 measures $138.04$ ms and
confirms the rejection. This empirical $10.08$ ms difference
between the graph-equivalent P1 and P2 artifacts becomes calibration evidence
for later candidates under the same context.

\begin{figure}[!t]
\centering
\includegraphics[width=0.98\linewidth]{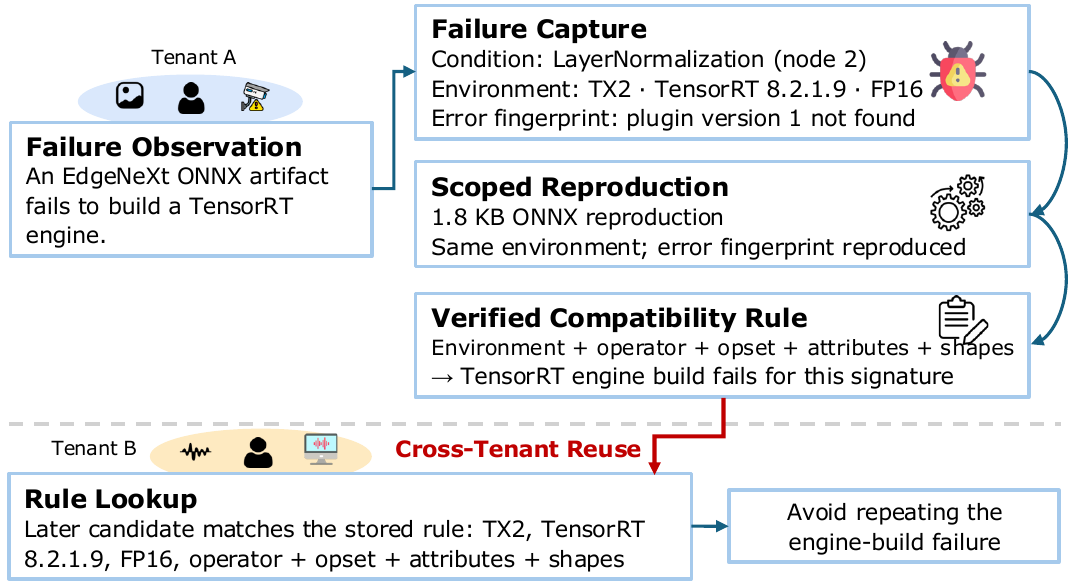}
\vspace{-6pt}
\caption{A reproduced device–runtime failure becomes a verified compatibility rule reused by a later tenant.}
\label{fig:failure-rule-trace}
\end{figure}

\subsection{Cross-Tenant Shared Services}
\label{sec:shared_services}

\noindent\textbf{Private trees, shared services.}
Each request keeps a private tree, while Cross-Tenant Shared Services support two forms of sharing.
(1) \textit{Pipeline sharing.}
Each candidate produces two ordered jobs: cloud training and
target-device execution. These jobs are sequential for one candidate, but can overlap across concurrent requests: one candidate can train while another executes. (2) \textit{Rule sharing.} P0 matches later candidates across tenants against verified compatibility rules, allowing reproduced failure conditions to prevent repeated failures and device work.

\noindent\textbf{Pipeline sharing.}
Training and device verification use different resource pools, so sequential
execution leaves one pool idle while the other works. Each job retains the dependency \emph{training and export $\rightarrow$ target-device verification}.
Across requests, the scheduler uses tenant-level round-robin to preserve fair
access to shared resources. After selecting a tenant, it runs that tenant’s
shortest ready job first. A shorter job releases its GPU or device sooner,
reducing head-of-line blocking and exposing the next pipeline stage earlier.
Because this ordering applies only within the selected tenant, round-robin scheduling
still controls progress across tenants. The tree prior breaks ties. Figure~\ref{fig:multi-tenant-runtime}
evaluates the resulting request-level queueing delay and completion time.

\noindent\textbf{Rule sharing: from failure to verified rule.}
Similar error text can come from different causes, while an operator name alone does not capture its attributes, tensor shapes, or runtime version. \name{}
therefore shares a failure only after reproducing its specific condition. When
artifact compilation, loading, or execution fails, \name{} records the runtime
environment, failure stage, normalized error, and the implicated artifact
condition, such as an ONNX header field or an operator with its attributes and
tensor shapes. It constructs a small reproducer and
runs it in the same device–runtime environment. Reproducing the same error
creates a verified compatibility rule that records the environment, reproduced
condition, verdict, and reproduction procedure. P0 consumes this rule under
the rejection conditions in \S~\ref{sec:multi_fidelity_verifier}. The same
verified condition can be reused when it recurs in later requests, including
those of other tenants. A runtime or driver change leaves the rule stale pending reproduction. Figure~\ref{fig:failure-rule-trace} traces one failure through
this path; Appendix~\ref{app:verifier-reproducibility} details reproduction and
rule matching.

\noindent\textbf{Tenant boundary.}
The scheduler reads tenant identity, resource and time estimates, target
device, and stage status. It does not inspect datasets, model semantics, or
mutation logic. Across requests, the service shares scheduling metadata and
sanitized verified compatibility rules. Data, generated code, model weights,
artifacts, and search state remain in the request workspace. Shared execution
keeps cloud and edge resources busy, while verified rules avoid repeated device
work without coupling private synthesis trees.

\vspace{-5pt}
\section{Evaluation}
Our evaluation seeks to answer the following questions:

\begin{enumerate}[label=Q\arabic*., wide=0pt, leftmargin=*, nosep]
\item How effectively does \name{} synthesize high-quality, SLO-feasible edge ML artifacts? (\S~\ref{subsec:eval_overall})

\item How do \name{}'s key designs and parameters affect synthesis quality and efficiency? (\S~\ref{subsec:eval_ablation}, \S~\ref{subsec:eval_sensitivity})

\item Can \name{} generalize to a challenging real-world task? (\S~\ref{subsec:eval_case_study}) Where is its capability boundary? (\S~\ref{subsec:boundary})

\end{enumerate}

\begin{figure*}[!t]
\centering
\includegraphics[width=0.95\textwidth]{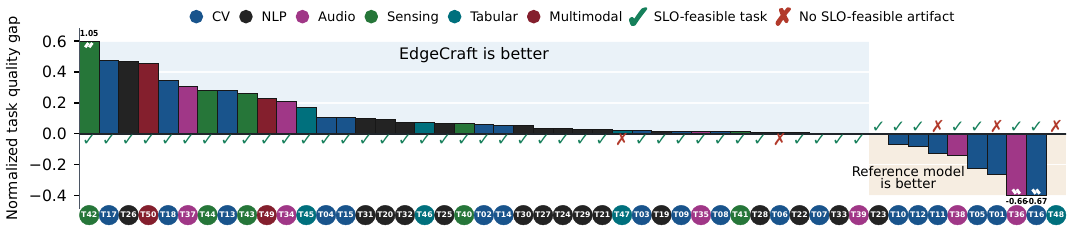}
\vspace{-5pt}
\caption{Normalized task quality gap of \name{} and reference models on 50 edge/mobile datasets.}
\label{fig:overall-quality}
\vspace{-8pt}
\end{figure*}

\begin{figure}[!t]
\centering
\includegraphics[width=\columnwidth]{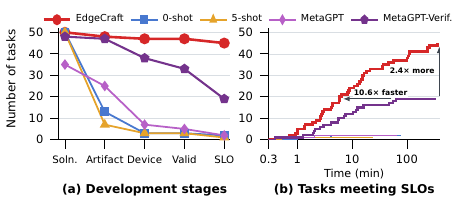}
\vspace{-16pt}
\caption{Agentic development across 50 tasks: (a) stage-wise task retention and (b) cumulative SLO-feasible tasks versus per-task time.}
\label{fig:automated-development}
\end{figure}

\noindent\textbf{Benchmark.} We construct a comprehensive benchmark encompassing 50 public datasets across six modalities: CV, NLP, Audio, Sensing \& Time Series, Tabular, and Multimodal (Table~\ref{tab:mcaas-benchmarks}, Appendix~\ref{app:Public_Benchmark}). These tasks represent essential IoT applications, such as smart homes and smart health, and are important to real-world evaluation in mobile computing. Moreover, we include a self-collected multimodal dataset to evaluate generalizability to a new task, as detailed in \S~\ref{subsec:eval_case_study}.
We carefully pair each dataset with target edge devices (Table~\ref{tab:device_runtime_summary}) based on computational complexity and realistic scenarios. For example, compute-intensive object detection is mapped to Jetson GPUs, tabular analysis is mapped to desktop CPUs, and lightweight sensing is assigned to Raspberry Pi 5.

To establish the evaluation basis, we require performance anchors for comparing task quality and on-device performance and for constructing consistent Reference-relative SLOs. We therefore select a Reference model for each public dataset. Each Reference is task-adapted and representative of established ML practices. It must also compile and execute successfully on its assigned device, enabling direct on-device performance comparisons. Table~\ref{tab:mcaas-benchmarks} in Appendix~\ref{app:Public_Benchmark} reports the Reference models and their performance; Appendix~\ref{app:benchmark-details} provides comprehensive evaluation details, including Reference selection and optimization, SLO construction, energy measurement, and dataset splits.

\begin{table}[t]
\centering
\footnotesize
\renewcommand{\arraystretch}{1.0}
\begin{tabularx}{\columnwidth}{l c l l}
\toprule
\textbf{Device} & \textbf{Abbr.} & \textbf{Compute Resources} & \textbf{Runtime} \\
\midrule
Jetson AGX Orin & Orin & 8-core Arm, Ampere GPU 32GB & P, O, T \\
Jetson Xavier NX & NX & 6-core Arm, Volta GPU 8GB & P, O, T \\
Jetson TX2 & TX2 & 4-core Arm, Pascal GPU 4GB & P, O, T \\
Raspberry Pi 5 & Pi5 & 4-core Cortex-A76 8GB & P, O, L \\
Desktop (CPU) & PC & Intel Ultra 5 235 & P, O, L \\
\bottomrule
\end{tabularx}
\vspace{5pt}
\caption{Evaluated hardware platforms. Runtime abbr.: P (PyTorch), O (ONNX), T (TensorRT), and L (LiteRT).}\label{tab:device_runtime_summary}
\vspace{-8pt}
\end{table}

\noindent\textbf{Baselines.}
Beyond the reference models, we compare \name{} with three automation paradigms. (1)~\textit{LLM zero-shot and five-shot generation} uses the same LLM backend and frozen task context as \name{}, but receives neither iterative repair nor verification results. (2)~\textit{MetaGPT Data Interpreter}~\cite{hong2024metagpt, Data_interpreter} is a popular, powerful, and representative agent framework that develops one candidate at a time through its native data-science workflow. We further implement \textit{MetaGPT-Verification}, which retains MetaGPT’s core workflow but trains each candidate on the server, deploys the resulting artifact to the target edge device, and returns its verification results to MetaGPT before the next revision. Comparing the two variants isolates the benefit of iterative target-device verification. (3)~\textit{AutoML and hardware-aware NAS frameworks} include FLAML~\cite{FLAML}, AutoML-Agent~\cite{Automl-agent}, AutoGluon~\cite{AutoGluon_agtabular,tang2024automm}, and Once-for-All (OFA)~\cite{cai2020onceforalltrainnetworkspecialize}. They represent highly optimized ML capabilities in their respective domains. Because each AutoML and NAS method supports only certain tasks and modalities, we evaluate it only on the datasets it supports. Moreover, we include AIDE’s tree-search policy~\cite{jiang2025aide} as a competitive tree-search baseline. We run it with the same settings and verifier as Constraint-Aware. Figure~\ref{fig:constraint-aware-expansion} measures how the choice among several search policies affects synthesis performance.

\noindent\textbf{Setup.} We host \name{} and all baselines on a server equipped with eight NVIDIA A6000 GPUs (48 GB), which connects to edge devices (Table \ref{tab:device_runtime_summary}) over a local area network. For the software stack, \name{}'s agentic backend is built upon LangChain~\cite{langchain2025} and LangGraph~\cite{langgraph2025}. GPT-5.4 serves as the default LLM engine. Each \name{} request and MetaGPT’s development loop may evaluate at most 24 executable candidates.

\subsection{Overall Performance}
\label{subsec:eval_overall}

\noindent\textbf{\name{} finds SLO-feasible artifacts on most edge/mobile ML tasks.}
Each request uses
$L_{\mathrm{SLO}}=0.8L_{\mathrm{Ref}}$ and, when energy is constrained,
$E_{\mathrm{SLO}}=0.8E_{\mathrm{Ref}}$, with Reference measurements taken on
the assigned device. This setting asks \name{} to improve task quality under tighter deployment limits. We define the normalized quality gap
as $g_Q=s_Q(Q_{\name{}}-Q_{\mathrm{Ref}})/|Q_{\mathrm{Ref}}|$, where $s_Q=1$
for higher-is-better metrics and $s_Q=-1$ otherwise; positive values favor
\name{}. Figure~\ref{fig:overall-quality} summarizes two complementary outcomes of EdgeCraft's search: the best task quality observed during development and whether the search reaches the SLO-feasible region. Across 50 tasks, EdgeCraft improves the best-observed quality over the Reference on 40 tasks (80\%) and finds at least one artifact satisfying all active SLOs on 45 tasks (90\%). Together, these results show that EdgeCraft explores competitive designs broadly and reaches deployable solutions on most tasks, even when the latency and energy limits are 20\% tighter than the corresponding Reference measurements. \S~\ref{subsec:eval_sensitivity} discusses the SLO sensitivity.

\noindent\textbf{\name{} outperforms agentic methods in SLO feasibility, artifact quality, and development time.} Figure~\ref{fig:automated-development}(a) follows each method through five cumulative stages: \emph{Solution}, all required loader, training, inference, and description files are present; \emph{Artifact}, training and export produce an artifact for the requested runtime; \emph{Device}, the artifact loads and completes inference on the assigned device; \emph{Valid}, its predictions and measurement output pass the task evaluator; and \emph{SLO}, the validated artifact satisfies every active task-quality, latency, and energy requirement.
The methods diverge sharply after generation: zero-shot, five-shot, and MetaGPT each produce SLO-feasible artifacts for only one or two tasks. Target-device verification raises MetaGPT-Verification to 19 feasible tasks, showing that verification matters; \name{} reaches 45, or 2.4$\times$ as many, showing that verification is more effective within \name{}'s structured design. The advantage also extends beyond feasibility. On the 16 tasks where both methods produce SLO-feasible artifacts, \name{} achieves higher quality on 15 (93.8\%) and ties on one, including 69.3\% versus 58.0\% accuracy on \inlinenlptag{T30} and 74.5\% versus 49.2\% mIoU on \inlinecvtag{T10}. Figure~\ref{fig:automated-development}(b) shows the same trend over time: \name{} reaches 19 SLO-feasible tasks in 5.9 minutes, whereas MetaGPT-Verification takes 62.8 minutes, 10.6$\times$ longer. Thus, \name{} produces feasible artifacts more broadly, improves their quality more consistently, and reaches them sooner.

To characterize the resources required by a complete development request, we measure one complete task from each of the six modalities across five hardware targets. Averaged across these requests, EdgeCraft uses 0.71 million GPT-5.4 tokens per request, compared with 1.61 million for MetaGPT-Verification, a 55.8\% reduction. EdgeCraft produces SLO-feasible artifacts for five of the six tasks, compared with three for MetaGPT-Verification, while using 1.22 versus 0.75 A6000 slot-hours per request.

\begin{figure}[!t]
\centering
\includegraphics[width=\columnwidth]{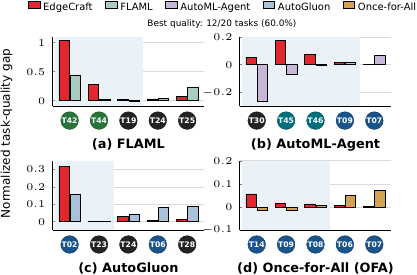}
\vspace{-15pt}
\caption{\name{} and AutoML/NAS baselines on their supported datasets.}
\label{fig:automl-supported-subset}
\end{figure}

\noindent\textbf{\name{} remains competitive in task quality against AutoML/NAS baselines.}
Figure~\ref{fig:automl-supported-subset} focuses exclusively on task quality. Each bar reports $g_Q=s_Q(Q_{\mathrm{method}}-Q_{\mathrm{Ref}})/|Q_{\mathrm{Ref}}|$, where a positive value favors the method. Across these 20 comparisons, \name{} achieves the highest task quality on 12 (60\%): three against FLAML, four against AutoML-Agent, two against AutoGluon, and three against Once-for-All. These results show that \name{} remains competitive in pure predictive quality even when compared with specialized optimization frameworks in the domains they are designed to support. In this experiment, each AutoML/NAS framework is evaluated on five datasets aligned with its native support and domain-optimized search space, allowing its specialized modeling capability to be fully exercised. Latency and energy SLO satisfaction do not enter this comparison.

\begin{figure}[!t]
\centering
\includegraphics[width=\columnwidth]{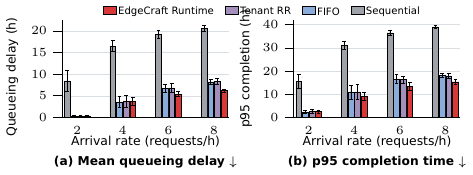}
\vspace{-20pt}
\caption{Multi-tenant performance at 2–8 requests/hour: (a) mean queueing delay and (b) p95 completion time (mean$\pm$s.d.).}
\label{fig:multi-tenant-runtime}
\end{figure}

\noindent\textbf{\name{}'s multi-tenant runtime reduces waiting and tail completion time.}
Figure~\ref{fig:multi-tenant-runtime} reports results for 64 request arrivals using measured candidate-job sequences and stage times from eight completed \name{} development runs. Each job enters the ready queue after its dependencies complete. All policies receive identical arrivals and work, so the measured differences come from their scheduling decisions. Sequential processes requests from beginning to end, FIFO schedules ready jobs in arrival order, and Tenant RR rotates across tenants. At 8 requests/hour, \name{} reduces mean queueing delay by 24.9\% and p95 completion time by 14.6\% relative to Tenant RR. These results show that the multi-tenant runtime provides an effective supporting substrate for concurrent requests.

\vspace{-5pt}
\subsection{Ablation Study}
\label{subsec:eval_ablation}

\begin{figure}[!t]
\centering
\includegraphics[width=\columnwidth]{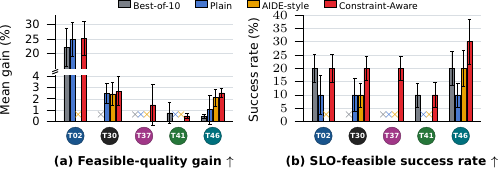}
\vspace{-15pt}
\caption{Under the same budget, Constraint-Aware leads in (a)
feasible-quality gain and (b) SLO-feasible incumbent-improvement rate;
$\times$: $\leq 0.05\%$.}
\label{fig:constraint-aware-expansion}
\end{figure}

\noindent\textbf{Verification-guided control produces better feasible artifacts under the same budget.}
Figure~\ref{fig:constraint-aware-expansion} compares four policies from the same four roots over ten additional candidates: independent Best-of-10, Plain without SLO-gap guidance, AIDE-style tree expansion~\cite{jiang2025aide}, and Constraint-Aware. Across five tasks spanning vision, language, audio, sensing, and tabular workloads, Constraint-Aware produces the largest feasible-quality gain on four and matches or exceeds every alternative in SLO-feasible improvement rate on all five. By using verified quality progress and SLO gaps to choose the next branch and revision, 10–30\% of its expansions improve the feasible incumbent, compared with 0–20\% for the other search strategies.

\begin{figure}[!t]
\centering
\includegraphics[width=\columnwidth]{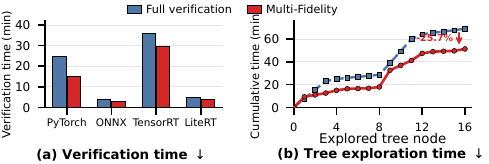}
\vspace{-16pt}
\caption{Multi-Fidelity cuts verification time by
17.6–39.7\% across the four runtimes in (a) and total exploration time by
25.7\% in (b). In all four runtimes, it selects the same artifact as full
verification.}
\label{fig:multifidelity-ablation}
\end{figure}

\noindent\textbf{Multi-Fidelity reduces verification cost while
preserving the final result.}
Figure~\ref{fig:multifidelity-ablation}(a) compares Multi-Fidelity with full
verification on 16 fixed candidates spanning PyTorch, ONNX Runtime, TensorRT,
and LiteRT. Multi-Fidelity cuts per-runtime verification time by 17.6–39.7\%
and selects the same artifact with the same task quality in every runtime.
Figure~\ref{fig:multifidelity-ablation}(b) accumulates this cost as the 16
candidates are explored: total time falls from 69.13 to 51.35 minutes
(25.7\%). The fast check applies only to graph-preserving
candidates whose device, runtime, and execution setting already have a
completed calibration; all others proceed to full verification. Across the
complete evaluation, all 60 early rejections are confirmed by full
verification.
Appendix~\ref{app:verifier-reproducibility} details the eligibility condition,
cold-start behavior, and safety analysis.

\begin{figure}[!t]
\centering
\includegraphics[width=\columnwidth]{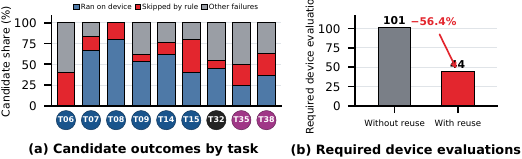}
\vspace{-16pt}
\caption{Verified Failure Reuse skips at least one candidate in each of the
nine evaluated tasks in (a) and reduces required device evaluations from 101
to 44 (56.4\%) across the full evaluation in (b).}
\label{fig:verified-failure-reuse}
\end{figure}

\noindent\textbf{Verified Failure Reuse avoids repeated device executions.}
EdgeCraft stores each reproduced failure as a rule scoped to its device and runtime; a candidate matching the same failure conditions is skipped before device evaluation. Figure~\ref{fig:verified-failure-reuse}(a) shows at least one such reuse in each of the nine evaluated TX2 tasks. The full evaluation contains 134 candidates from 15 datasets spanning five modalities, four devices, and three runtimes. All 50 verified rules are reused at least once and collectively match 57 candidates (42.5\%). Figure~\ref{fig:verified-failure-reuse}(b) shows that these matches reduce required device evaluations from 101 to 44 (56.4\%). Device execution confirms the predicted failure for all 57 matched candidates. Appendix~\ref{app:verifier-reproducibility} reports and discusses the rule composition, cross-dataset reuse, and how workload shifts affect future matches.

\subsection{Sensitivity Analysis}
\label{subsec:eval_sensitivity}

\begin{figure}[!t]
\centering
\includegraphics[width=\columnwidth]{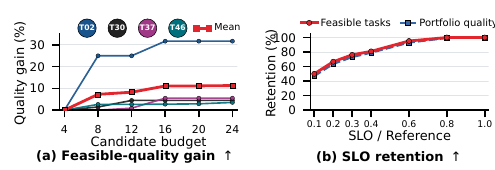}
\vspace{-15pt}
\caption{Search and SLO sensitivity: (a) feasible-quality gain with
budget; (b) feasible-task and task-normalized quality retention for the fixed
$1.0\times$ cohort.}
\label{fig:sensitivity-analysis}
\end{figure}

\noindent\textbf{\name{} remains effective across search budgets and LLM backends.}
Figure~\ref{fig:sensitivity-analysis}(a) shows feasible-quality gain as a function of the search budget. From 4 to 24 candidates, the best verified SLO-feasible
quality improves on every task, averaging 11.3\% and reaching 31.7\%. On
\inlineaudiotag{T37}, the same \name{} workflow achieves verified
SLO-feasible accuracy of 67.2\%, 52.2\%, 58.9\%, 55.0\%, and 79.4\% with
GPT-5.4, GPT-5.2, DeepSeek-V4-Pro, Gemini-3.1-Pro, and Claude-Opus-4.8,
respectively. All five backends produce a verified SLO-feasible artifact, and the resulting accuracies broadly align with the relative performance of the corresponding LLMs.

\noindent\textbf{\name{} retains broad coverage as SLOs tighten.}
Figure~\ref{fig:sensitivity-analysis}(b) jointly tightens each task's latency
and applicable energy limits for the 42 tasks with complete measurements. At
$0.4\times$ and $0.3\times$, 34 and 32 tasks remain feasible, retaining 79.1\%
and 73.1\% of the aggregate task-normalized quality. Most surviving tasks keep
their best $1.0\times$ artifact: 29/34 at $0.4\times$ and 25/32 at
$0.3\times$. The remaining cases show how the search adapts. At $0.3\times$,
\inlinecvtag{T17} moves from MobileNetV3 to a tiny depthwise crowd regressor in
TensorRT, whereas \inlinesensetag{T41} uses a lightweight ONNX model whose
accuracy changes from 95.5\% to 93.0\%.

\subsection{Generalizability}
\label{subsec:eval_case_study}

\noindent \textbf{SEN (self-collected)\footnote{The study received institutional review board approval; informed consent was obtained, participation was voluntary, and all records were de-identified.}.}
Beyond the public benchmark, we include the SEN dataset for a generalizability study. It contains 304~MB of multimodal biosignals collected from 29 students with special educational needs (SEN) using Empatica E4 wristbands. Its four sensing modalities are accelerometry (ACC), blood-volume pulse (BVP), galvanic skin response (GSR), and skin temperature. We use SEN for emotion recognition (happy, sad, neutral). This task presents representative challenges for lightweight wearable sensing. We ensure that raw SEN data remain local; the LLM receives only the dataset schema, modality summaries, and candidate-level verification results.
We use a sample-level split and a subject-disjoint split whose six test subjects never appear in training. We select competitive baselines covering three families. (1) \emph{Feature engineering + RF} maps each modality to a fixed feature vector, concatenates the vectors, and fits a Random Forest (RF) classifier~\cite{breiman2001randomforests}. Expert+RF uses modality-specific physiological and temporal descriptors tailored to ACC, BVP, GSR, and temperature. General+RF extracts, per signal, mean, standard deviation, maximum, skewness, kurtosis, and quartiles in the time domain, together with spectral centroid, spread, mean and peak frequency, and spectral quartiles. Kats+RF uses 40 time-series characteristics spanning distribution, trend, seasonality, and nonlinear dynamics~\cite{facebook2021kats}; Catch22+RF uses 22 compact, diverse time-series characteristics~\cite{lubba2019catch22}. (2) \emph{Learned features and models} include MiniRocket~\cite{dempster2021minirocket}, which uses convolutional features, and NormWear~\cite{luo2026normwear}, a large end-to-end multimodal foundation model for wearable tasks. (3) \emph{AutoML and agent-based baselines} include AutoGluon and MetaGPT.

\begin{table}[!t]
\centering
\scriptsize
\renewcommand{\arraystretch}{1.0}
\setlength{\tabcolsep}{5.4pt}
\begin{tabularx}{\columnwidth}{@{}*{4}{>{\centering\arraybackslash}X}@{}}
\toprule
\textbf{Method} & \textbf{Sample-level $\uparrow$} & \textbf{Cross-subject $\uparrow$} & \textbf{Avg. $\uparrow$} \\
\midrule
Expert+RF & $0.8241\!\pm\!0.0010$ & $0.5477\!\pm\!0.0048$ & 0.6859 \\
General+RF & $0.8945\!\pm\!0.0024$ & $0.5280\!\pm\!0.0037$ & 0.7113 \\
Kats+RF & $0.7803\!\pm\!0.0008$ & $0.5595\!\pm\!0.0013$ & 0.6699 \\
Catch22+RF & $0.5744\!\pm\!0.0053$ & $0.5322\!\pm\!0.0004$ & 0.5533 \\
MiniRocket & $0.7883\!\pm\!0.0006$ & {\bfseries\boldmath $0.5697\!\pm\!0.0044$} & 0.6790 \\
NormWear & $0.8414\!\pm\!0.0168$ & $0.5293\!\pm\!0.0170$ & 0.6854 \\
AutoGluon & $0.8766\!\pm\!0.0123$ & $0.5148\!\pm\!0.0013$ & 0.6957 \\
MetaGPT & $0.6849\!\pm\!0.0326$ & $0.5224\!\pm\!0.0447$ & 0.6037 \\
\textbf{EdgeCraft} & {\bfseries\boldmath $0.8953\!\pm\!0.0180$} & $0.5683\!\pm\!0.0092$ & \textbf{0.7318} \\
\midrule
\multicolumn{4}{@{}c@{}}{%
\begin{tabularx}{\columnwidth}{@{}>{\centering\arraybackslash}X>{\centering\arraybackslash}X>{\centering\arraybackslash}X@{}}
\textit{Deployment metric} & \textbf{Sample-level} & \textbf{Cross-subject} \\
p95 latency / SLO (ms) & $0.251\!\pm\!0.065$ / 0.4 & $0.352\!\pm\!0.015$ / 0.4 \\
Energy / SLO (mJ) & $1.052\!\pm\!0.560$ / 2.8 & $1.581\!\pm\!0.591$ / 2.8 \\
EdgeCraft LLM tokens & $1.139\!\pm\!0.088$M & $0.772\!\pm\!0.570$M \\
\end{tabularx}} \\
\bottomrule
\end{tabularx}%
\vspace{4pt}
\caption{SEN AUROC and Pi~5 deployment results.}
\label{tab:sen-generalizability}
\vspace{-8pt}
\end{table}

We use the same Raspberry Pi~5 SLO for both splits: p95 latency $\leq 0.4$~ms and energy $\leq 2.8$~mJ per inference. As shown in Table~\ref{tab:sen-generalizability}, \name{} achieves the highest sample-level mean AUROC of $0.8953\pm0.0180$, demonstrating its effectiveness in exploring high-quality sensing models. The subject-disjoint setting is substantially harder: its six test subjects never appear during training, and even NormWear, a large foundation model for multimodal wearable sensing, reaches only 0.5293 AUROC. Under this shift, \name{} achieves $0.5683\pm0.0092$, essentially matching the strongest task-specific baseline, MiniRocket, at $0.5697\pm0.0044$, while outperforming all other evaluated baselines. Across all three runs, every selected artifact also satisfies the shared Raspberry Pi~5 SLOs. Thus, \name{} achieves the highest predictive quality when subject-specific patterns are available and remains competitive on entirely unseen subjects.

\subsection{Capability Boundary}\label{subsec:boundary}

\noindent \textbf{Reference quality and SLO feasibility.}
Figure~\ref{fig:overall-quality} reports task-quality gain and SLO feasibility
across all 50 tasks. The
highest-quality candidates trail their respective References on nine tasks, and
\inlinetabtag{T48} produces no valid artifact, yielding 40 quality wins overall.
The largest gaps occur on \inlinecvtag{T16} and \inlineaudiotag{T36} ($-0.67$
and $-0.66$), where the task-specific FastReID and HuBERT References retain
substantially more quality than the compact deployable candidates; the other
seven gaps range from $-0.01$ to $-0.26$. Five tasks—\inlinecvtag{T01},
\inlinecvtag{T06}, \inlinecvtag{T11}, \inlinetabtag{T47}, and
\inlinetabtag{T48}—are marked with a cross because their searches find no
SLO-feasible artifact, leaving 45 tasks with at least one feasible artifact.

\noindent\textbf{Coverage degrades under stringent SLOs.}
At $0.1\times$, only 21 of the 42 tasks remain feasible: 17 use ONNX Runtime, two use TensorRT, and two use PyTorch. They include linear text models, MFCC/log-mel audio networks, and tiny depthwise or reduced-input vision models. These survivors retain 93.8\% mean quality (100\% median) relative to their best feasible artifacts at $1.0\times$. However, after accounting for the tasks that lose feasibility, joint coverage–quality retention drops to 46.9\%. This exposes a clear boundary: extreme SLOs favor aggressively compressible tasks and substantially narrow overall coverage.

\vspace{-5pt}
\section{Discussion}

\noindent\textbf{Solution optimality and scope boundary.}
\name{} targets workloads where compact models, data representations, or
runtime optimizations can preserve quality under physical SLOs. Its boundary
arises when quality depends on a high-capacity, task-specific backbone that
cannot fit these SLOs. \name{} also does not guarantee a global optimum. SLO gaps guide revisions, and it returns the highest-quality artifact fully verified within the candidate budget. Data collection, labeling, and task definition remain outside the scope of \name{}'s automation pipeline.

\noindent\textbf{Privacy.}
Each user provides an intent, dataset, target device, and SLOs, and receives the resulting artifact and measurements. In multi-user mode, requests are authenticated and kept in separate workspaces, and users can access only their own tasks and artifacts through the service. Shared components process each request but do not share user data or artifacts across users; only limited scheduling information, sanitized measurement summaries, and verified compatibility rules are reused.

\noindent\textbf{Public-benchmark exposure.}
A potential concern is that LLMs may have encountered information about public datasets during pretraining. This exposure may benefit both \name{} and LLM-driven baselines and bias comparisons. However, it cannot be ruled out because the LLM pretraining corpora are outside our control. We reduce its influence by evaluating 50 tasks across diverse modalities and concrete task–device–runtime settings, so the evaluation does not hinge on a narrow benchmark set. We further include the self-collected SEN dataset, which was unavailable during pretraining.
Performance on SEN demonstrates that \name{} can  generalize beyond publicly available benchmarks.

\vspace{-5pt}
\section{Related Work}\label{sec:related_work}

\noindent\textbf{Managed low-code ML platforms.}
Industrial low-code and managed ML platforms, including
EasyDL~\cite{baidu_easydl_2026}, SageMaker AI~\cite{aws_sagemaker_ai_2025}, and Vertex AI Vision~\cite{google_vertex_ai_vision_2026}, reduce engineering effort through graphical workflows, hosted training, model catalogs, and deployment services. They make established ML pipelines easier to build and operate, while users still choose the task formulation, data representation, model family, export path, and target runtime. \name{} automates these coupled choices from high-level intent and returns the complete ML artifact.

\noindent\textbf{AutoML, HPO, and NAS.}
AutoML automates model selection and training through HPO and NAS.
Multi-fidelity methods such as Hyperband, BOHB, and ASHA allocate partial
resource budgets and progressively promote promising
configurations~\cite{li2018hyperbandnovelbanditbasedapproach,falkner2018bohbrobustefficienthyperparameter,li2020massivelyparallelhyperparametertuning};
hardware-aware NAS further incorporates predicted or measured device
costs~\cite{tan2019mnasnetplatformawareneuralarchitecture,cai2019proxylessnasdirectneuralarchitecture,cai2020onceforalltrainnetworkspecialize,zhang2021nnmeter,10.1145/3560905.3568520}.
These approaches use a prescribed search space and a common candidate
interface, allowing lower-cost measurements to be compared across
configurations. LLM-based AutoML agents broaden this workflow to agentic ML development for data science tasks~\cite{Ds-agent,Automl-agent,Automl-gpt}. \name{} instead searches complete ML programs whose data
pipelines, model graphs, training procedures, export paths, and runtimes may
differ, and whose failures can occur before or during deployment. It calibrates
fast checks against earlier full verification and reuses reproduced device--runtime failures as scoped
rejection rules. The resulting deployment evidence guides full verification
and directs later search.

\noindent\textbf{Coding agents.}
LLMs can plan, generate, execute, and repair code across multi-stage
workflows~\cite{karpathy_autoresearch_2026,novikov2025alphaevolve,jiang2025aide,hong2025datainterpreter,hong2024metagpt,romera2024funsearch}.
Agentic skills further package reusable domain knowledge and
procedures~\cite{liu2026harnessingllmagentsskill}. \name{} uses this general
orchestration capability for edge ML development, while grounding each branch
in measured task quality, latency, energy, and runtime status on the assigned
device. A task is complete only when its deployed artifact satisfies the
requirements, and the same evidence guides branch expansion, revision,
pruning, and acceptance.

\noindent\textbf{LLM-driven IoT program synthesis.} LLMs also show strong capabilities in IoT applications.
TaskSense translates user intent into executable sensor toolchains for heterogeneous sensor
systems~\cite{liu2025tasksense}; AutoIOT synthesizes AIoT application
code~\cite{shen2025autoiot}; and EmbedGenius automates embedded IoT software
development~\cite{yang2024embedgenius}. Together, they broaden the programming
interface for embedded systems. Although those systems may involve ML components, they remain program-centric rather than model-centric. \name{} focuses on automated edge-model development and deployment.

\vspace{-5pt}
\section{Conclusion}
We propose \name{}, a verification-grounded system that couples constraint-aware LLM synthesis with progressive target-device verification, reusable verified compatibility rules, and multi-tenant execution. Across 50 tasks, \name{} combines broad development capability and competitive quality with lower verification time and token use.

\newpage

\bibliographystyle{ACM-Reference-Format}
\bibliography{reference}

@article{taneja2023artificial,
  title={Artificial intelligence: Implications for the agri-food sector},
  author={Taneja, Akriti and Nair, Gayathri and Joshi, Manisha and Sharma, Somesh and Sharma, Surabhi and Jambrak, Anet Rezek and Rosell{\'o}-Soto, Elena and Barba, Francisco J and Castagnini, Juan M and Leksawasdi, Noppol and others},
  journal={Agronomy},
  volume={13},
  number={5},
  pages={1397},
  year={2023},
  publisher={MDPI}
}

@ARTICLE{wearable_sensing,
  author={Mukhopadhyay, Subhas Chandra},
  journal={IEEE Sensors Journal},
  title={Wearable Sensors for Human Activity Monitoring: A Review},
  year={2015},
  volume={15},
  number={3},
  pages={1321-1330},

  doi={10.1109/JSEN.2014.2370945}}

@inproceedings{srivastava2017safety,
  title={Safety and security in smart cities using artificial intelligence—A review},
  author={Srivastava, Shweta and Bisht, Aditya and Narayan, Neetu},
  booktitle={2017 7th international conference on cloud computing, data science \& engineering-confluence},
  pages={130--133},
  year={2017},
  organization={IEEE}
}

@misc{he2026egologegocentricfinegraineddaily,
      title={EgoLog: Ego-Centric Fine-Grained Daily Log with Ubiquitous Wearables},
      author={Lixing He and Bufang Yang and Di Duan and Zhenyu Yan and Guoliang Xing},
      year={2026},
      eprint={2504.02624},
      archivePrefix={arXiv},
      primaryClass={cs.HC},
      url={https://arxiv.org/abs/2504.02624},
}

@article{qi2025alzheimer,
  title={Alzheimer’s disease digital biomarkers multidimensional landscape and AI model scoping review},
  author={Qi, Wenhao and Zhu, Xiaohong and Wang, Bin and Shi, Yankai and Dong, Chaoqun and Shen, Shiying and Li, Jiaqi and Zhang, Kun and He, Yunfan and Zhao, Mengjiao and others},
  journal={npj Digital Medicine},
  volume={8},
  number={1},
  pages={366},
  year={2025},
  publisher={Nature Publishing Group UK London}
}

@article{singh2023edge,
  title={Edge AI: a survey},
  author={Singh, Raghubir and Gill, Sukhpal Singh},
  journal={Internet of Things and Cyber-Physical Systems},
  volume={3},
  pages={71--92},
  year={2023},
  publisher={Elsevier}
}

@article{zhou2019edge,
  title={Edge intelligence: Paving the last mile of artificial intelligence with edge computing},
  author={Zhou, Zhi and Chen, Xu and Li, En and Zeng, Liekang and Luo, Ke and Zhang, Junshan},
  journal={Proceedings of the IEEE},
  volume={107},
  number={8},
  pages={1738--1762},
  year={2019},
  publisher={IEEE}
}

@article{david2020globalWheat,
         title={Global Wheat Head Detection (GWHD) Dataset: A Large and Diverse Dataset of High-Resolution RGB-Labelled Images to Develop and Benchmark Wheat Head Detection Methods},
         author={David, Etienne and Madec, Simon and Sadeghi-Tehran, Pouria and Aasen, Helge and Zheng, Bangyou and Liu, Shouyang and Kirchgessner, Norbert and Ishikawa, Goro and Nagasawa, Koichi and Badhon, Minhajul and others},
         journal={arXiv preprint arXiv:2005.02162},
         year={2020}
}

@dataset{Jocher_Ultralytics_Datasets_2025_HomeObjects_3K,
    author = {Jocher, Glenn and Rizwan, Muhammad},
    license = {AGPL-3.0},
    month = {May},
    title = {Ultralytics Datasets: HomeObjects-3K Detection Dataset},
    url = {https://docs.ultralytics.com/datasets/detect/homeobjects-3k/},
    version = {1.0.0},
    year = {2025}
}

@TECHREPORT{Krizhevsky09learningmultiple_cifar10_cifar100,
            author={Alex Krizhevsky},
            title={Learning multiple layers of features from tiny images},
            institution={},
            year={2009}
}

@article{speechcommandsv2,
   author = { {Warden}, P.},
    title = "{Speech Commands: A Dataset for Limited-Vocabulary Speech Recognition}",
  journal = {ArXiv e-prints},
  archivePrefix = "arXiv",
  eprint = {1804.03209},
  primaryClass = "cs.CL",

    year = 2018,
    month = apr,
    url = {https://arxiv.org/abs/1804.03209},
}

@article{fei2007learning_Caltech_101,
  title={Learning generative visual models from few training examples: An incremental Bayesian approach tested on 101 object categories},
  author={Fei-Fei, Li and Fergus, Rob and Perona, Pietro},
  journal={Computer vision and Image understanding},
  volume={106},
  number={1},
  pages={59--70},
  year={2007},
  publisher={Elsevier}
}

@misc{crack-bphdr_dataset,
    title = { crack Dataset },
    type = { Open Source Dataset },
    author = { University },
    url = { https://universe.roboflow.com/university-bswxt/crack-bphdr },
    year = { 2022 },
    month = { dec },
    note = { visited on 2024-01-23 },
}

@misc{everingham2010pascal_voc,
      title={The PASCAL Visual Object Classes (VOC) Challenge},
      author={Mark Everingham and Luc Van Gool and Christopher K. I. Williams and John Winn and Andrew Zisserman},
      year={2010},
      eprint={0909.5206},
      archivePrefix={arXiv},
      primaryClass={cs.CV}
}

@inproceedings{khosla2011fgvc,
  title={Novel dataset for Fine-Grained Image Categorization},
  author={Aditya Khosla and Nityananda Jayadevaprakash and Bangpeng Yao and Li Fei-Fei},
  booktitle={First Workshop on Fine-Grained Visual Categorization (FGVC), IEEE Conference on Computer Vision and Pattern Recognition (CVPR)},
  year={2011}
}

@misc{ultralytics_dog_pose,
  author       = {{Ultralytics}},
  title        = {Dog-Pose Estimation Dataset},
  year         = {2024},
  howpublished = {Ultralytics Datasets},
  url          = {https://docs.ultralytics.com/datasets/pose/dog-pose/},
  note         = {Accessed: 2026-08-06}
}

@article{Geiger2013IJRR_KITTI,
  author = {Andreas Geiger and Philip Lenz and Christoph Stiller and Raquel Urtasun},
  title = {Vision meets Robotics: The KITTI Dataset},
  journal = {International Journal of Robotics Research (IJRR)},
  year = {2013}
}

@article{Automl-agent,
  title={Automl-agent: A multi-agent llm framework for full-pipeline automl},
  author={Trirat, Patara and Jeong, Wonyong and Hwang, Sung Ju},
  journal={arXiv preprint arXiv:2410.02958},
  year={2024}
}

@article{Automl-gpt,
  title={Automl-gpt: Automatic machine learning with gpt},
  author={Zhang, Shujian and Gong, Chengyue and Wu, Lemeng and Liu, Xingchao and Zhou, Mingyuan},
  journal={arXiv preprint arXiv:2305.02499},
  year={2023}
}

@article{Ds-agent,
  title={Ds-agent: Automated data science by empowering large language models with case-based reasoning},
  author={Guo, Siyuan and Deng, Cheng and Wen, Ying and Chen, Hechang and Chang, Yi and Wang, Jun},
  journal={arXiv preprint arXiv:2402.17453},
  year={2024}
}

@inproceedings{Data_interpreter,
  title={Data interpreter: An llm agent for data science},
  author={Hong, Sirui and Lin, Yizhang and Liu, Bang and Liu, Bangbang and Wu, Binhao and Zhang, Ceyao and Li, Danyang and Chen, Jiaqi and Zhang, Jiayi and Wang, Jinlin and others},
  booktitle={Findings of the Association for Computational Linguistics: ACL 2025},
  pages={19796--19821},
  year={2025}
}

@article{he2021automl,
  title={AutoML: A survey of the state-of-the-art},
  author={He, Xin and Zhao, Kaiyong and Chu, Xiaowen},
  journal={Knowledge-based systems},
  volume={212},
  pages={106622},
  year={2021},
  publisher={Elsevier}
}

@inproceedings{redmon2016you,
  title={You only look once: Unified, real-time object detection},
  author={Redmon, Joseph and Divvala, Santosh and Girshick, Ross and Farhadi, Ali},
  booktitle={Proceedings of the IEEE conference on computer vision and pattern recognition},
  pages={779--788},
  year={2016}
}

@misc{ultralytics2026,
  author = {{Ultralytics Inc.}},
  title = {Ultralytics | Revolutionizing the World of Vision AI},
  howpublished = {\url{https://www.ultralytics.com/}},
  year = {2026},
  note = {Accessed: 2026-01-29}
}

@misc{onnx_github,
  author       = {{ONNX Community}},
  title        = {{GitHub - onnx/onnx: Open standard for machine learning interoperability}},
  year         = {2026},
  howpublished = {\url{https://github.com/onnx/onnx}},
  note         = {GitHub repository. Accessed: 2026-04-13}
}

@misc{tensorrt_homepage,
  author       = {{NVIDIA}},
  title        = {{NVIDIA TensorRT}},
  year         = {2026},
  howpublished = {\url{https://developer.nvidia.com/tensorrt}},
  note         = {Official NVIDIA Developer page. Accessed: 2026-04-13}
}

@misc{karpathy_autoresearch_2026,
  author       = {Andrej Karpathy},
  title        = {{karpathy/autoresearch}},
  year         = {2026},
  howpublished = {\url{https://github.com/karpathy/autoresearch}},
  note         = {GitHub repository. Accessed: 2026-04-13}
}

@misc{MobileNet,
      title={MobileNets: Efficient Convolutional Neural Networks for Mobile Vision Applications},
      author={Andrew G. Howard and Menglong Zhu and Bo Chen and Dmitry Kalenichenko and Weijun Wang and Tobias Weyand and Marco Andreetto and Hartwig Adam},
      year={2017},
      eprint={1704.04861},
      archivePrefix={arXiv},
      primaryClass={cs.CV},
      url={https://arxiv.org/abs/1704.04861},
}

@misc{MobileNetV2,
      title={MobileNetV2: Inverted Residuals and Linear Bottlenecks},
      author={Mark Sandler and Andrew Howard and Menglong Zhu and Andrey Zhmoginov and Liang-Chieh Chen},
      year={2019},
      eprint={1801.04381},
      archivePrefix={arXiv},
      primaryClass={cs.CV},
      url={https://arxiv.org/abs/1801.04381},
}

@misc{MobileNetV3,
      title={Searching for MobileNetV3},
      author={Andrew Howard and Mark Sandler and Grace Chu and Liang-Chieh Chen and Bo Chen and Mingxing Tan and Weijun Wang and Yukun Zhu and Ruoming Pang and Vijay Vasudevan and Quoc V. Le and Hartwig Adam},
      year={2019},
      eprint={1905.02244},
      archivePrefix={arXiv},
      primaryClass={cs.CV},
      url={https://arxiv.org/abs/1905.02244},
}

@misc{raspberrypi,
  author = {{Raspberry Pi Foundation}},
  title = {Raspberry Pi},
  howpublished = {\url{https://www.raspberrypi.org/}},
  year = {2026},
  note = {Accessed: 2026-01-29}
}

@misc{nvidia_jetson_2026,
  author = {{NVIDIA Corporation}},
  title = {The Ultimate Platform for Physical AI and Robotics},
  howpublished = {\url{https://www.nvidia.com/en-us/autonomous-machines/embedded-systems/}},
  year = {2026},
  note = {Accessed: 2026-01-29}
}

@misc{pytorch_homepage_2026,
  author       = {{PyTorch Foundation}},
  title        = {{PyTorch}},
  year         = {2026},
  howpublished = {\url{https://pytorch.org/}},
  note         = {Accessed: 2026-01-29}
}

@article{novikov2025alphaevolve,
  title={Alphaevolve: A coding agent for scientific and algorithmic discovery},
  author={Novikov, Alexander and V{\~u}, Ng{\^a}n and Eisenberger, Marvin and Dupont, Emilien and Huang, Po-Sen and Wagner, Adam Zsolt and Shirobokov, Sergey and Kozlovskii, Borislav and Ruiz, Francisco JR and Mehrabian, Abbas and others},
  journal={arXiv preprint arXiv:2506.13131},
  year={2025}
}

@inproceedings{shen2025autoiot,
  title={Autoiot: Llm-driven automated natural language programming for aiot applications},
  author={Shen, Leming and Yang, Qiang and Zheng, Yuanqing and Li, Mo},
  booktitle={Proceedings of the 31st Annual International Conference on Mobile Computing and Networking},
  pages={468--482},
  year={2025}
}

@inproceedings{liu2025tasksense,
  title={TaskSense: A Translation-like Approach for Tasking Heterogeneous Sensor Systems with LLMs},
  author={Liu, Kaiwei and Yang, Bufang and Xu, Lilin and Guo, Yunqi and Xing, Guoliang and Shuai, Xian and Ren, Xiaozhe and Jiang, Xin and Yan, Zhenyu},
  booktitle={Proceedings of the 23rd ACM Conference on Embedded Networked Sensor Systems},
  pages={213--225},
  year={2025}
}

@article{jiang2025aide,
  title={Aide: Ai-driven exploration in the space of code},
  author={Jiang, Zhengyao and Schmidt, Dominik and Srikanth, Dhruv and Xu, Dixing and Kaplan, Ian and Jacenko, Deniss and Wu, Yuxiang},
  journal={arXiv preprint arXiv:2502.13138},
  year={2025}
}

@inproceedings{hong2025datainterpreter,
  title={Data interpreter: An llm agent for data science},
  author={Hong, Sirui and Lin, Yizhang and Liu, Bang and Liu, Bangbang and Wu, Binhao and Zhang, Ceyao and Li, Danyang and Chen, Jiaqi and Zhang, Jiayi and Wang, Jinlin and others},
  booktitle={Findings of the Association for Computational Linguistics: ACL 2025},
  pages={19796--19821},
  year={2025}
}

@inproceedings{IoT-Brain,
author = {Zhou, Zhaomeng and Zhang, Lan and Wang, Junyang and Yuan, Mu},
title = {IoT-Brain: Grounding LLMs for Semantic-Spatial Sensor Scheduling},
year = {2025},
isbn = {9798400719813},
publisher = {Association for Computing Machinery},
address = {New York, NY, USA},
url = {https://doi.org/10.1145/3737904.3768530},
doi = {10.1145/3737904.3768530},

booktitle = {Proceedings of the 2025 ACM Workshop on Access Networks with Artificial Intelligence},
pages = {6–10},
numpages = {5},

location = {
},
series = {ANAI '25}
}

@misc{baidu_easydl_2026,
  author       = {{Baidu}},
  title        = {{EasyDL}},
  year         = {2026},
  howpublished = {\url{https://ai.baidu.com/easydl/}},
  note         = {Official website. Accessed: 2026-04-13}
}

@misc{aws_sagemaker_ai_2025,
  author       = {{Amazon Web Services}},
  title        = {{Machine Learning Service - Amazon SageMaker AI}},
  year         = {2025},
  howpublished = {\url{https://aws.amazon.com/pm/sagemaker/}},
  note         = {Official website. Accessed: 2026-04-13}
}

@misc{google_vertex_ai_vision_2026,
  author       = {{Google Cloud}},
  title        = {{Vertex AI Vision}},
  year         = {2026},
  howpublished = {\url{https://docs.cloud.google.com/vision-ai/docs}},
  note         = {Official documentation page, last updated 2026-04-11. Accessed: 2026-04-13}
}

@article{yang2024embedgenius,
  title={Embedgenius: Towards automated software development for generic embedded iot systems},
  author={Yang, Huanqi and Li, Mingzhe and Han, Mingda and Li, Zhenjiang and Xu, Weitao},
  journal={arXiv preprint arXiv:2412.09058},
  year={2024}
}

@misc{h2o_python_2016,
  author       = {{H2O.ai}},
  title        = {Python Interface for H2O, Python Module Version 3.10.0.8},
  year         = {2016},
  month        = oct,
  url          = {https://github.com/h2oai/h2o-3}
}

@article{AutoGluon_agtabular,
  title={AutoGluon-Tabular: Robust and Accurate AutoML for Structured Data},
  author={Erickson, Nick and Mueller, Jonas and Shirkov, Alexander and Zhang, Hang and Larroy, Pedro and Li, Mu and Smola, Alexander},
  journal={arXiv preprint arXiv:2003.06505},
  year={2020}
}

@article{breiman2001randomforests,
  author  = {Breiman, Leo},
  title   = {Random Forests},
  journal = {Machine Learning},
  volume  = {45},
  number  = {1},
  pages   = {5--32},
  year    = {2001},
  doi     = {10.1023/A:1010933404324}
}

@software{facebook2021kats,
author = {Jiang, Xiaodong and Srivastava, Sudeep and Chatterjee, Sourav and Yu, Yang and Handler, Jeffrey and Zhang, Peiyi and Bopardikar, Rohan and Li, Dawei and Lin, Yanjun and Thakore, Uttam and Brundage, Michael and Holt, Ginger and Komurlu, Caner and Nagalla, Rakshita and Wang, Zhichao and Sun, Hechao and Gao, Peng and Cheung, Wei and Gao, Jun and Wang, Qi and Guerard, Marius and Kazemi, Morteza and Chen, Yulin and Zhou, Chong and Lee, Sean and Laptev, Nikolay and Levendovszky, Tihamér and Taylor, Jake and Qian, Huijun and Zhang, Jian and Shoydokova, Aida and Singh, Trisha and Zhu, Chengjun and Baz, Zeynep and Bergmeir, Christoph and Yu, Di and Koylan, Ahmet and Jiang, Kun and Temiyasathit, Ploy and Yurtbay, Emre},
license = {MIT License},
month = {3},
title = {{Kats}},
url = {https://github.com/facebookresearch/Kats},
version = {0.2.0},
year = {2022}
}

@article{lubba2019catch22,
  author  = {Lubba, Carl H. and Sethi, Sarab S. and Knaute, Philip and Schultz, Simon R. and Fulcher, Ben D. and Jones, Nick S.},
  title   = {catch22: {CAnonical} Time-series {CHaracteristics}},
  journal = {Data Mining and Knowledge Discovery},
  volume  = {33},
  number  = {6},
  pages   = {1821--1852},
  year    = {2019},
  doi     = {10.1007/s10618-019-00647-x}
}

@inproceedings{dempster2021minirocket,
  author    = {Dempster, Angus and Schmidt, Daniel F. and Webb, Geoffrey I.},
  title     = {{MINIROCKET}: A Very Fast (Almost) Deterministic Transform for Time Series Classification},
  booktitle = {Proceedings of the 27th ACM SIGKDD Conference on Knowledge Discovery and Data Mining},
  publisher = {Association for Computing Machinery},
  address   = {New York, NY, USA},
  pages     = {248--257},
  year      = {2021},
  doi       = {10.1145/3447548.3467231}
}

@article{luo2026normwear,
  author  = {Luo, Yunfei and Chen, Yuliang and Salekin, Asif and Rahman, Tauhidur},
  title   = {Toward Foundation Model for Multivariate Wearable Sensing of Physiological Signals},
  journal = {ACM Transactions on Computing for Healthcare},
  volume  = {7},
  number  = {3},
  articleno = {43},
  numpages = {43},
  year    = {2026},
  doi     = {10.1145/3803808}
}

@misc{yang2025proagentharnessingondemandsensory,
      title={ProAgent: Harnessing On-Demand Sensory Contexts for Proactive LLM Agent Systems},
      author={Bufang Yang and Lilin Xu and Liekang Zeng and Yunqi Guo and Siyang Jiang and Wenrui Lu and Kaiwei Liu and Hancheng Xiang and Xiaofan Jiang and Guoliang Xing and Zhenyu Yan},
      year={2025},
      eprint={2512.06721},
      archivePrefix={arXiv},
      primaryClass={cs.AI},
      url={https://arxiv.org/abs/2512.06721},
}

@software{torchvision2016,
    title        = {TorchVision: PyTorch's Computer Vision library},
    author       = {TorchVision maintainers and contributors},
    year         = 2016,
    journal      = {GitHub repository},
    publisher    = {GitHub},
    howpublished = {\url{https://github.com/pytorch/vision}}
}

@misc{rw2019timm,
  author = {Ross Wightman},
  title = {PyTorch Image Models},
  year = {2019},
  publisher = {GitHub},
  journal = {GitHub repository},
  doi = {10.5281/zenodo.4414861},
  howpublished = {\url{https://github.com/rwightman/pytorch-image-models}}
}

@inproceedings{hong2024metagpt,
      title={Meta{GPT}: Meta Programming for A Multi-Agent Collaborative Framework},
      author={Sirui Hong and Mingchen Zhuge and Jonathan Chen and Xiawu Zheng and Yuheng Cheng and Jinlin Wang and Ceyao Zhang and Zili Wang and Steven Ka Shing Yau and Zijuan Lin and Liyang Zhou and Chenyu Ran and Lingfeng Xiao and Chenglin Wu and J{\"u}rgen Schmidhuber},
      booktitle={The Twelfth International Conference on Learning Representations},
      year={2024},
      url={https://openreview.net/forum?id=VtmBAGCN7o}
}

@misc{openai_gpt5_chat,
  author       = {{OpenAI}},
  title        = {{GPT-5 Chat Model}},
  howpublished = {\url{https://openai.com/zh-Hans-CN/index/introducing-gpt-5/}},
  year         = {2025},
  note         = {Accessed: 2026-04-28}
}

@inproceedings{zhang2021nnmeter,
  title={Nn-meter: Towards accurate latency prediction of deep-learning model inference on diverse edge devices},
  author={Zhang, Li Lyna and Han, Shihao and Wei, Jianyu and Zheng, Ningxin and Cao, Ting and Yang, Yuqing and Liu, Yunxin},
  booktitle={Proceedings of the 19th Annual International Conference on Mobile Systems, Applications, and Services},
  pages={81--93},
  year={2021}
}

@misc{ASAP,
      title={ASAP: Architecture Search, Anneal and Prune},
      author={Asaf Noy and Niv Nayman and Tal Ridnik and Nadav Zamir and Sivan Doveh and Itamar Friedman and Raja Giryes and Lihi Zelnik-Manor},
      year={2019},
      eprint={1904.04123},
      archivePrefix={arXiv},
      primaryClass={stat.ML},
      url={https://arxiv.org/abs/1904.04123},
}

@misc{liu2026instmeter,
      title={InstMeter: An Instruction-Level Method to Predict Energy and Latency of DL Model Inference on MCUs},
      author={Hao Liu and Qing Wang and Marco Zuniga},
      year={2026},
      eprint={2603.04134},
      archivePrefix={arXiv},
      primaryClass={cs.LG},
      url={https://arxiv.org/abs/2603.04134},
}

@article{romera2024funsearch,
  title={Mathematical discoveries from program search with large language models},
  author={Romera-Paredes, Bernardino and Barekatain, Mohammadamin and Novikov, Alexander and Balog, Matej and Kumar, M Pawan and Dupont, Emilien and Ruiz, Francisco JR and Ellenberg, Jordan S and Wang, Pengming and Fawzi, Omar and others},
  journal={Nature},
  volume={625},
  number={7995},
  pages={468--475},
  year={2024},
  publisher={Nature Publishing Group UK London}
}

@misc{langchain2025,
  author       = {LangChain AI},
  title        = {{LangChain}: The Agent Engineering Platform},
  year         = {2025},
  publisher    = {GitHub},
  howpublished = {\url{https://github.com/langchain-ai/langchain}},
  note         = {A framework for building agents and LLM-powered applications
                  with interoperable components and third-party integrations},
  license      = {MIT},
  url          = {https://docs.langchain.com/langchain/}
}

@misc{langgraph2025,
  author       = {LangChain AI},
  title        = {{LangGraph}: A Low-Level Orchestration Framework and Runtime
                  for Building Stateful Agents},
  year         = {2025},
  publisher    = {LangChain Inc.},
  howpublished = {\url{https://docs.langchain.com/oss/python/langgraph/overview}},
  note         = {Provides durable execution, streaming, human-in-the-loop,
                  and persistence for long-running agent workflows.
                  Inspired by Pregel and Apache Beam}
}

@online{paperswithcode,
  author       = {{Papers with Code}},
  title        = {Papers with Code},
  year         = {2026},
  url          = {https://paperswithcode.com/},
  note         = {Website accessed: 2026-07-01}
}

@online{kaggle,
  author       = {{Kaggle}},
  title        = {Kaggle: The World's AI Proving Ground
},
  year         = {2026},
  url          = {https://www.kaggle.com/},
  note         = {Website accessed: 2026-07-01}
}

@misc{FLAML,
      title={FLAML: A Fast and Lightweight AutoML Library},
      author={Chi Wang and Qingyun Wu and Markus Weimer and Erkang Zhu},
      year={2021},
      eprint={1911.04706},
      archivePrefix={arXiv},
      primaryClass={cs.LG},
      url={https://arxiv.org/abs/1911.04706},
}

@ARTICLE{VisDrone,
  author={Zhu, Pengfei and Wen, Longyin and Du, Dawei and Bian, Xiao and Fan, Heng and Hu, Qinghua and Ling, Haibin},
  journal={IEEE Transactions on Pattern Analysis and Machine Intelligence},
  title={Detection and Tracking Meet Drones Challenge},
  year={2021},
  volume={},
  number={},
  pages={1-1},
  doi={10.1109/TPAMI.2021.3119563}
}

@INPROCEEDINGS{liu2018ano_pred,
        author={W. Liu and W. Luo, D. Lian and S. Gao},
        booktitle={2018 IEEE Conference on Computer Vision and Pattern Recognition (CVPR)},
        title={Future Frame Prediction for Anomaly Detection -- A New Baseline},
        year={2018}
}

@article{telelogs,
  title={{Reasoning Language Models for Root Cause Analysis in 5G Wireless Networks}},
  author={Mohamed Sana and Nicola Piovesan and Antonio De Domenico and Yibin Kang and Haozhe Zhang and Merouane Debbah and Fadhel Ayed},
  year={2025},
  eprint={[arXiv preprint arXiv:2507.21974]},
  url={https://arxiv.org/abs/2507.21974}
}

@misc{loghub1,
      title={Loghub: A Large Collection of System Log Datasets for AI-driven Log Analytics},
      author={Jieming Zhu and Shilin He and Pinjia He and Jinyang Liu and Michael R. Lyu},
      year={2023},
      eprint={2008.06448},
      archivePrefix={arXiv},
      primaryClass={cs.SE},
      url={https://arxiv.org/abs/2008.06448},
}

@misc{loghub2,
      title={A Large-Scale Evaluation for Log Parsing Techniques: How Far Are We?},
      author={Zhihan Jiang and Jinyang Liu and Junjie Huang and Yichen Li and Yintong Huo and Jiazhen Gu and Zhuangbin Chen and Jieming Zhu and Michael R. Lyu},
      year={2024},
      eprint={2308.10828},
      archivePrefix={arXiv},
      primaryClass={cs.SE},
      url={https://arxiv.org/abs/2308.10828},
}

@misc{SLURP,
      title={SLURP: A Spoken Language Understanding Resource Package},
      author={Emanuele Bastianelli and Andrea Vanzo and Pawel Swietojanski and Verena Rieser},
      year={2020},
      eprint={2011.13205},
      archivePrefix={arXiv},
      primaryClass={cs.CL},
      url={https://arxiv.org/abs/2011.13205},
}

@article{MVTecAD1,
  title={The MVTec Anomaly Detection Dataset: A Comprehensive Real-World Dataset for Unsupervised Anomaly Detection},
  author={Bergmann, Paul and Batzner, Kilian and Fauser, Michael and Sattlegger, David and Steger, Carsten},
  journal={International Journal of Computer Vision},
  volume={129},
  number={4},
  pages={1038--1059},
  year={2021},
  doi={10.1007/s11263-020-01400-4}
}

@inproceedings{MVTecAD2,
  title={MVTec AD — A Comprehensive Real-World Dataset for Unsupervised Anomaly Detection},
  author={Bergmann, Paul and Fauser, Michael and Sattlegger, David and Steger, Carsten},
  booktitle={IEEE/CVF Conference on Computer Vision and Pattern Recognition (CVPR)},
  pages={9584--9592},
  year={2019},
  doi={10.1109/CVPR.2019.00982}
}

@inproceedings{xia2018dota,
  author    = {Xia, Gui-Song and Bai, Xiang and Ding, Jian and Zhu, Zhen and
               Belongie, Serge and Luo, Jiebo and Datcu, Mihai and Pelillo,
               Marcello and Zhang, Liangpei},
  title     = {{DOTA}: A Large-Scale Dataset for Object Detection in Aerial Images},
  booktitle = {Proceedings of the IEEE Conference on Computer Vision and Pattern Recognition},
  year      = {2018},
  pages     = {3974--3983},
  doi       = {10.1109/CVPR.2018.00418}
}

@article{song2013neu,
  author  = {Song, Kechen and Yan, Yunhui},
  title   = {A Noise Robust Method Based on Completed Local Binary Patterns for Hot-Rolled Steel Strip Surface Defects},
  journal = {Applied Surface Science},
  volume  = {285},
  pages   = {858--864},
  year    = {2013},
  doi     = {10.1016/j.apsusc.2013.09.002},
  url     = {https://faculty.neu.edu.cn/songkechen/zh_CN/zdylm/263270/list/index.htm}
}

@inproceedings{casanueva2020banking77,
  author    = {Casanueva, I{\~n}igo and Tem\v{c}inas, Tadas and Gerz, Daniela and
               Henderson, Matthew and Vuli{\'c}, Ivan},
  title     = {Efficient Intent Detection with Dual Sentence Encoders},
  booktitle = {Proceedings of the 2nd Workshop on Natural Language Processing for Conversational AI},
  year      = {2020},
  pages     = {38--45},
  url       = {https://aclanthology.org/2020.nlp4convai-1.5/}
}

@misc{cai2020onceforalltrainnetworkspecialize,
      title={Once-for-All: Train One Network and Specialize it for Efficient Deployment},
      author={Han Cai and Chuang Gan and Tianzhe Wang and Zhekai Zhang and Song Han},
      year={2020},
      eprint={1908.09791},
      archivePrefix={arXiv},
      primaryClass={cs.LG},
      url={https://arxiv.org/abs/1908.09791},
}

@misc{yang2018netadaptplatformawareneuralnetwork,
      title={NetAdapt: Platform-Aware Neural Network Adaptation for Mobile Applications},
      author={Tien-Ju Yang and Andrew Howard and Bo Chen and Xiao Zhang and Alec Go and Mark Sandler and Vivienne Sze and Hartwig Adam},
      year={2018},
      eprint={1804.03230},
      archivePrefix={arXiv},
      primaryClass={cs.CV},
      url={https://arxiv.org/abs/1804.03230},
}

@misc{tan2019mnasnetplatformawareneuralarchitecture,
      title={MnasNet: Platform-Aware Neural Architecture Search for Mobile},
      author={Mingxing Tan and Bo Chen and Ruoming Pang and Vijay Vasudevan and Mark Sandler and Andrew Howard and Quoc V. Le},
      year={2019},
      eprint={1807.11626},
      archivePrefix={arXiv},
      primaryClass={cs.CV},
      url={https://arxiv.org/abs/1807.11626},
}

@inproceedings{PieBridge,
author = {Yin, Wangsong and Xu, Daliang and Huang, Gang and Zhang, Ying and Wei, Shiyun and Xu, Mengwei and Liu, Xuanzhe},
title = {PieBridge: Fast and Parameter-Efficient On-Device Training via Proxy Networks},
year = {2024},
isbn = {9798400706974},
publisher = {Association for Computing Machinery},
address = {New York, NY, USA},
url = {https://doi.org/10.1145/3666025.3699327},
doi = {10.1145/3666025.3699327},

booktitle = {Proceedings of the 22nd ACM Conference on Embedded Networked Sensor Systems},
pages = {126–140},
numpages = {15},

location = {Hangzhou, China},
series = {SenSys '24}
}

@article{tang2024automm,
  title={AutoGluon-Multimodal (AutoMM): Supercharging Multimodal AutoML with Foundation Models},
  author={Tang, Zhiqiang and Fang, Haoyang and Zhou, Su and Yang, Taojiannan and Zhong, Zihan and Hu, Tony and Kirchhoff, Katrin and Karypis, George},
  journal={arXiv preprint arXiv:2404.16233},
  year={2024}
}

@misc{wolf2020huggingfacestransformersstateoftheartnatural,
      title={HuggingFace's Transformers: State-of-the-art Natural Language Processing},
      author={Thomas Wolf and Lysandre Debut and Victor Sanh and Julien Chaumond and Clement Delangue and Anthony Moi and Pierric Cistac and Tim Rault and Rémi Louf and Morgan Funtowicz and Joe Davison and Sam Shleifer and Patrick von Platen and Clara Ma and Yacine Jernite and Julien Plu and Canwen Xu and Teven Le Scao and Sylvain Gugger and Mariama Drame and Quentin Lhoest and Alexander M. Rush},
      year={2020},
      eprint={1910.03771},
      archivePrefix={arXiv},
      primaryClass={cs.CL},
      url={https://arxiv.org/abs/1910.03771},
}

@misc{liu2026harnessingllmagentsskill,
      title={Harnessing LLM Agents with Skill Programs},
      author={Hongjun Liu and Yifei Ming and Shafiq Joty and Chen Zhao},
      year={2026},
      eprint={2605.17734},
      archivePrefix={arXiv},
      primaryClass={cs.AI},
      url={https://arxiv.org/abs/2605.17734},
}

@inproceedings{clinc150,
    title = "An Evaluation Dataset for Intent Classification and Out-of-Scope Prediction",
    author = "Larson, Stefan  and
      Mahendran, Anish  and
      Peper, Joseph J.  and
      Clarke, Christopher  and
      Lee, Andrew  and
      Hill, Parker  and
      Kummerfeld, Jonathan K.  and
      Leach, Kevin  and
      Laurenzano, Michael A.  and
      Tang, Lingjia  and
      Mars, Jason",
    booktitle = "Proceedings of the 2019 Conference on Empirical Methods in Natural Language Processing and the 9th International Joint Conference on Natural Language Processing (EMNLP-IJCNLP)",
    year = "2019",
    url = "https://www.aclweb.org/anthology/D19-1131"
}

@inproceedings{Stallkamp-IJCNN-2011-gtsrb,
    author = {Johannes Stallkamp and Marc Schlipsing and Jan Salmen and Christian Igel},
    booktitle = {IEEE International Joint Conference on Neural Networks},
    title = {The {G}erman {T}raffic {S}ign {R}ecognition {B}enchmark: A multi-class classification competition},
    year = {2011},
    pages = {1453--1460}
}

@misc{zhang2016ag-news,
      title={Character-level Convolutional Networks for Text Classification},
      author={Xiang Zhang and Junbo Zhao and Yann LeCun},
      year={2016},
      eprint={1509.01626},
      archivePrefix={arXiv},
      primaryClass={cs.LG},
      url={https://arxiv.org/abs/1509.01626},
}

@article{Eidinger2014AgeAG,
    author = "Eidinger, Eran and Enbar, Roee and Hassner, Tal",
    title = "Age and Gender Estimation of Unfiltered Faces",
    journal = "IEEE Transactions on Information Forensics and Security",
    year = "2014",
    volume = "9",
    pages = "2170-2179"
}

@INPROCEEDINGS{7553002,
  author={Liu, Xinchen and Liu, Wu and Ma, Huadong and Fu, Huiyuan},
  booktitle={2016 IEEE International Conference on Multimedia and Expo (ICME)},
  title={Large-scale vehicle re-identification in urban surveillance videos},
  year={2016},
  volume={},
  number={},
  pages={1-6},

  doi={10.1109/ICME.2016.7553002}}

@dataset{gautam_2019_3355823,
  author       = {Gautam},
  title        = {E commerce text dataset},
  month        = jul,
  year         = 2019,
  publisher    = {Zenodo},
  version      = {version - 2},
  doi          = {10.5281/zenodo.3355823},
  url          = {https://doi.org/10.5281/zenodo.3355823},
}

@inproceedings{barbieri2020tweeteval,
title={{TweetEval:Unified Benchmark and Comparative Evaluation for Tweet Classification}},
author={Barbieri, Francesco and Camacho-Collados, Jose and Espinosa-Anke, Luis and Neves, Leonardo},
booktitle={Proceedings of Findings of EMNLP},
year={2020}
}

@misc{sms_spam_collection_228,
  author       = {Almeida, Tiago and Hidalgo, Jos},
  title        = {{SMS Spam Collection}},
  year         = {2011},
  howpublished = {UCI Machine Learning Repository},
  note         = {{DOI}: https://doi.org/10.24432/C5CC84}
}

@inproceedings{socher-etal-2013-recursive,
    title = "Recursive Deep Models for Semantic Compositionality Over a Sentiment Treebank",
    author = "Socher, Richard  and
      Perelygin, Alex  and
      Wu, Jean  and
      Chuang, Jason  and
      Manning, Christopher D.  and
      Ng, Andrew  and
      Potts, Christopher",
    booktitle = "Proceedings of the 2013 Conference on Empirical Methods in Natural Language Processing",
    month = oct,
    year = "2013",
    address = "Seattle, Washington, USA",
    publisher = "Association for Computational Linguistics",
    url = "https://www.aclweb.org/anthology/D13-1170",
    pages = "1631--1642",
}

@inproceedings{wang2019glue,
  title={{GLUE}: A Multi-Task Benchmark and Analysis Platform for Natural Language Understanding},
  author={Wang, Alex and Singh, Amanpreet and Michael, Julian and Hill, Felix and Levy, Omer and Bowman, Samuel R.},
  note={In the Proceedings of ICLR.},
  year={2019}
}

@misc{demszky2020goemotionsdatasetfinegrainedemotions,
      title={GoEmotions: A Dataset of Fine-Grained Emotions},
      author={Dorottya Demszky and Dana Movshovitz-Attias and Jeongwoo Ko and Alan Cowen and Gaurav Nemade and Sujith Ravi},
      year={2020},
      eprint={2005.00547},
      archivePrefix={arXiv},
      primaryClass={cs.CL},
      url={https://arxiv.org/abs/2005.00547},
}

@inproceedings{motionsense,
author = {Malekzadeh, Mohammad and Clegg, Richard G. and Cavallaro, Andrea and Haddadi, Hamed},
title = {Mobile Sensor Data Anonymization},
booktitle = {Proceedings of the International Conference on Internet of Things Design and Implementation},
series = {IoTDI '19},
year = {2019},
isbn = {978-1-4503-6283-2},
location = {Montreal, Quebec, Canada},
pages = {49--58},
numpages = {10},
url = {http://doi.acm.org/10.1145/3302505.3310068},
doi = {10.1145/3302505.3310068},
acmid = {3310068},
publisher = {ACM},
address = {New York, NY, USA},

}

@misc{radford2021learningtransferablevisualmodels,
      title={Learning Transferable Visual Models From Natural Language Supervision},
      author={Alec Radford and Jong Wook Kim and Chris Hallacy and Aditya Ramesh and Gabriel Goh and Sandhini Agarwal and Girish Sastry and Amanda Askell and Pamela Mishkin and Jack Clark and Gretchen Krueger and Ilya Sutskever},
      year={2021},
      eprint={2103.00020},
      archivePrefix={arXiv},
      primaryClass={cs.CV},
      url={https://arxiv.org/abs/2103.00020},
}

@misc{mrmars1010_banana_quality_dataset,
  author       = {MRMARS1010},
  title        = {Banana Quality Dataset},
  publisher    = {Kaggle},
  year         = {2024},
  url          = {https://www.kaggle.com/datasets/mrmars1010/banana-quality-dataset},
  note         = {Accessed: 2026-08-06}
}

@inproceedings{panayotov2015librispeech,
  title={Librispeech: an ASR corpus based on public domain audio books},
  author={Panayotov, Vassil and Chen, Guoguo and Povey, Daniel and Khudanpur, Sanjeev},
  booktitle={Acoustics, Speech and Signal Processing (ICASSP), 2015 IEEE International Conference on},
  pages={5206--5210},
  year={2015},
  organization={IEEE}
}

@dataset{qu2020wildblueberry,
  author    = {Qu, Hongchun and Obsie, Efrem and Drummond, Frank},
  title     = {Data for: Wild blueberry yield prediction using a combination of computer simulation and machine learning algorithms},
  year      = {2020},
  publisher = {Mendeley Data},
  version   = {V1},
  doi       = {10.17632/p5hvjzsvn8.1},
  url       = {https://doi.org/10.17632/p5hvjzsvn8.1}
}

@misc{hhar,
  author       = {Blunck, Henrik and Bhattacharya, Sourav and Prentow, Thor and Kjrgaard, Mikkel and Dey, Anind},
  title        = {{Heterogeneity Activity Recognition}},
  year         = {2015},
  howpublished = {UCI Machine Learning Repository},
  doi          = {10.24432/C5689X},
  url          = {https://doi.org/10.24432/C5689X}
}

@inproceedings{clark2019boolq,
  title     = {BoolQ: Exploring the Surprising Difficulty of Natural Yes/No Questions},
  author    = {Clark, Christopher and Lee, Kenton and Chang, Ming-Wei and Kwiatkowski, Tom and Collins, Michael and Toutanova, Kristina},
  booktitle = {Proceedings of NAACL},
  year      = {2019}
}

@article{livingstone2018ravdess,
  author  = {Livingstone, Steven R. and Russo, Frank A.},
  title   = {The Ryerson Audio-Visual Database of Emotional Speech and Song ({RAVDESS}): A dynamic, multimodal set of facial and vocal expressions in North American English},
  journal = {PLOS ONE},
  volume  = {13},
  number  = {5},
  pages   = {e0196391},
  year    = {2018},
  doi     = {10.1371/journal.pone.0196391},
  url     = {https://doi.org/10.1371/journal.pone.0196391}
}

@inproceedings{piczak2015dataset,
  title = {{ESC}: {Dataset} for {Environmental Sound Classification}},
  author = {Piczak, Karol J.},
  booktitle = {Proceedings of the 23rd {Annual ACM Conference} on {Multimedia}},
  date = {2015-10-13},
  year = {2015},
  url = {http://dl.acm.org/citation.cfm?doid=2733373.2806390},
  doi = {10.1145/2733373.2806390},
  location = {{Brisbane, Australia}},
  isbn = {978-1-4503-3459-4},
  publisher = {{ACM Press}},
  pages = {1015--1018}
}

@article{hodosh2013framing,
  title={Framing image description as a ranking task: Data, models and evaluation metrics},
  author={Hodosh, Micah and Young, Peter and Hockenmaier, Julia},
  journal={Journal of Artificial Intelligence Research},
  volume={47},
  pages={853--899},
  year={2013}
}

@misc{feedback_prize_ell_2022,
  author       = {{The Learning Agency Lab}},
  title        = {Feedback Prize - English Language Learning},
  publisher    = {Kaggle},
  year         = {2022},
  url          = {https://www.kaggle.com/competitions/feedback-prize-english-language-learning/data},
  note         = {Accessed: 2026-08-06}
}

@misc{bhojani_airline_reviews,
  author       = {Bhojani, Juhi},
  title        = {Airline Reviews},
  year         = {2024},
  publisher    = {Kaggle},
  url          = {https://www.kaggle.com/datasets/juhibhojani/airline-reviews},
  note         = {Accessed: 2026-08-06}
}

@misc{harode_webmd_drug_reviews_dataset,
  author       = {Harode, Rohan},
  title        = {WebMD Drug Reviews Dataset},
  publisher    = {Kaggle},
  year         = {2020},
  url          = {https://www.kaggle.com/datasets/rohanharode07/webmd-drug-reviews-dataset},
  note         = {Accessed: 2026-08-06}
}

@misc{abalone_1,
  author       = {Nash, Warwick and Sellers, Tracy and Talbot, Simon and Cawthorn, Andrew and Ford, Wes},
  title        = {{Abalone}},
  year         = {1994},
  howpublished = {UCI Machine Learning Repository},
  doi          = {10.24432/C55C7W},
  url          = {https://doi.org/10.24432/C55C7W}
}

@misc{drossos2019clothoaudiocaptioningdataset,
      title={Clotho: An Audio Captioning Dataset},
      author={Konstantinos Drossos and Samuel Lipping and Tuomas Virtanen},
      year={2019},
      eprint={1910.09387},
      archivePrefix={arXiv},
      primaryClass={cs.SD},
      url={https://arxiv.org/abs/1910.09387},
}

@article{liu2016heart_sound_database,
  author  = {Liu, Chengyu and Springer, David and Li, Qiao and Moody, Benjamin and Juan, Ricardo A. and Chorro, Francisco J. and Castells, Francisco and Roig, Jose M. and Silva, Ikaro and Johnson, Alistair E. W. and Syed, Zeeshan and Schmidt, Samuel E. and Papadaniil, Christina D. and Hadjileontiadis, Leontios J. and Naseri, Hosein and Moukadem, Ali and Dieterlen, Alain and Brandt, Christian and Tang, Hong and Samieinasab, Maryam and Samieinasab, Mohammad Reza and Sameni, Reza and Mark, Roger G. and Clifford, Gari D.},
  title   = {An open access database for the evaluation of heart sound algorithms},
  journal = {Physiological Measurement},
  year    = {2016},
  month   = {dec},
  volume  = {37},
  number  = {12},
  pages   = {2181--2213},
  doi     = {10.1088/0967-3334/37/12/2181},
  url     = {https://doi.org/10.1088/0967-3334/37/12/2181}
}

@misc{ultralytics_hand_keypoints,
  author       = {{Ultralytics}},
  title        = {Hand Keypoints Pose Estimation Dataset},
  year         = {2024},
  howpublished = {Ultralytics Datasets},
  url          = {https://docs.ultralytics.com/datasets/pose/hand-keypoints/},
  note         = {Accessed: 2026-08-06}
}

@inproceedings{zhang2016shanghaitech,
  author    = {Zhang, Yingying and Zhou, Desen and Chen, Siqin and Gao, Shenghua and Ma, Yi},
  title     = {Single-Image Crowd Counting via Multi-Column Convolutional Neural Network},
  booktitle = {Proceedings of the IEEE Conference on Computer Vision and Pattern Recognition},
  pages     = {589--597},
  year      = {2016},
  url       = {https://openaccess.thecvf.com/content_cvpr_2016/html/Zhang_Single-Image_Crowd_Counting_CVPR_2016_paper.html}
}

@inproceedings{lugosch2019fsc,
  author    = {Lugosch, Loren and Ravanelli, Mirco and Ignoto, Patrick and Tomar, Vikrant Singh and Bengio, Yoshua},
  title     = {Speech Model Pre-Training for End-to-End Spoken Language Understanding},
  booktitle = {Interspeech 2019},
  pages     = {814--818},
  year      = {2019},
  doi       = {10.21437/Interspeech.2019-2396},
  url       = {https://www.isca-archive.org/interspeech_2019/lugosch19_interspeech.html}
}

@dataset{benazir2024fsc,
  author    = {Benazir, Afsara},
  title     = {Fluent Speech Commands Dataset},
  publisher = {Zenodo},
  year      = {2024},
  doi       = {10.5281/zenodo.11106540},
  url       = {https://doi.org/10.5281/zenodo.11106540}
}

@inproceedings{large2018ethanol,
  author    = {Large, James and Kemsley, E. Kate and Wellner, Nikolaus and Goodall, Ian and Bagnall, Anthony},
  title     = {Detecting Forged Alcohol Non-invasively Through Vibrational Spectroscopy and Machine Learning},
  booktitle = {Advances in Knowledge Discovery and Data Mining},
  pages     = {298--309},
  publisher = {Springer},
  year      = {2018},
  doi       = {10.1007/978-3-319-93034-3_24},
  url       = {https://doi.org/10.1007/978-3-319-93034-3_24}
}

@article{birbaumer1999spelling,
  author  = {Birbaumer, Niels and Ghanayim, N. and Hinterberger, T. and Iversen, I. and Kotchoubey, B. and K{\"u}bler, A. and Perelmouter, J. and Taub, E. and Flor, H.},
  title   = {A Spelling Device for the Paralysed},
  journal = {Nature},
  volume  = {398},
  number  = {6725},
  pages   = {297--298},
  year    = {1999},
  doi     = {10.1038/18581},
  url     = {https://doi.org/10.1038/18581}
}

@misc{cevik_software_defect_prediction,
  author       = {Cevik, Mustafa},
  title        = {Software Defect Prediction},
  publisher    = {Kaggle},
  year         = {2019},
  url          = {https://www.kaggle.com/datasets/semustafacevik/software-defect-prediction},
  note         = {Version 1; accessed: 2026-08-06}
}

@dataset{clotho_v2_1,
  author    = {Drossos, Konstantinos and Lipping, Samuel and Virtanen, Tuomas},
  title     = {Clotho Dataset},
  publisher = {Zenodo},
  version   = {2.1},
  year      = {2021},
  url       = {https://zenodo.org/records/4783391}
}

@misc{li2018hyperbandnovelbanditbasedapproach,
      title={Hyperband: A Novel Bandit-Based Approach to Hyperparameter Optimization},
      author={Lisha Li and Kevin Jamieson and Giulia DeSalvo and Afshin Rostamizadeh and Ameet Talwalkar},
      year={2018},
      eprint={1603.06560},
      archivePrefix={arXiv},
      primaryClass={cs.LG},
      url={https://arxiv.org/abs/1603.06560},
}

@misc{li2020massivelyparallelhyperparametertuning,
      title={A System for Massively Parallel Hyperparameter Tuning},
      author={Liam Li and Kevin Jamieson and Afshin Rostamizadeh and Ekaterina Gonina and Moritz Hardt and Benjamin Recht and Ameet Talwalkar},
      year={2020},
      eprint={1810.05934},
      archivePrefix={arXiv},
      primaryClass={cs.LG},
      url={https://arxiv.org/abs/1810.05934},
}

@misc{yin2025elasticondevicellmservice,
      title={Elastic On-Device LLM Service},
      author={Wangsong Yin and Rongjie Yi and Daliang Xu and Gang Huang and Mengwei Xu and Xuanzhe Liu},
      year={2025},
      eprint={2409.09071},
      archivePrefix={arXiv},
      primaryClass={cs.DC},
      url={https://arxiv.org/abs/2409.09071},
}

@misc{falkner2018bohbrobustefficienthyperparameter,
      title={BOHB: Robust and Efficient Hyperparameter Optimization at Scale},
      author={Stefan Falkner and Aaron Klein and Frank Hutter},
      year={2018},
      eprint={1807.01774},
      archivePrefix={arXiv},
      primaryClass={cs.LG},
      url={https://arxiv.org/abs/1807.01774},
}

@misc{cai2019proxylessnasdirectneuralarchitecture,
      title={ProxylessNAS: Direct Neural Architecture Search on Target Task and Hardware},
      author={Han Cai and Ligeng Zhu and Song Han},
      year={2019},
      eprint={1812.00332},
      archivePrefix={arXiv},
      primaryClass={cs.LG},
      url={https://arxiv.org/abs/1812.00332},
}

@inproceedings{10.1145/3450268.3453520,
author = {Zhao, Zhihe and Wang, Kai and Ling, Neiwen and Xing, Guoliang},
title = {EdgeML: An AutoML Framework for Real-Time Deep Learning on the Edge},
year = {2021},
isbn = {9781450383547},
publisher = {Association for Computing Machinery},
address = {New York, NY, USA},
url = {https://doi.org/10.1145/3450268.3453520},
doi = {10.1145/3450268.3453520},

booktitle = {Proceedings of the International Conference on Internet-of-Things Design and Implementation},
pages = {133–144},
numpages = {12},

location = {Charlottesvle, VA, USA},
series = {IoTDI '21}
}

@misc{panchal2026mosaicruntimeefficientmultiagentembodied,
      title={Mosaic: Runtime-Efficient Multi-Agent Embodied Planning},
      author={Kunjal Panchal and Saayan Mitra and Sunav Choudhary and Victor Bursztyn and Somdeb Sarkhel and Hui Guan},
      year={2026},
      eprint={2607.09603},
      archivePrefix={arXiv},
      primaryClass={cs.MA},
      url={https://arxiv.org/abs/2607.09603},
}

@inproceedings{10.1145/3560905.3568520,
author = {Ling, Neiwen and Huang, Xuan and Zhao, Zhihe and Guan, Nan and Yan, Zhenyu and Xing, Guoliang},
title = {BlastNet: Exploiting Duo-Blocks for Cross-Processor Real-Time DNN Inference},
year = {2023},
isbn = {9781450398862},
publisher = {Association for Computing Machinery},
address = {New York, NY, USA},
url = {https://doi.org/10.1145/3560905.3568520},
doi = {10.1145/3560905.3568520},

booktitle = {Proceedings of the 20th ACM Conference on Embedded Networked Sensor Systems},
pages = {91–105},
numpages = {15},

location = {Boston, Massachusetts},
series = {SenSys '22}
}

@misc{zhang2025open3dvqabenchmarkcomprehensivespatial,
      title={Open3D-VQA: A Benchmark for Comprehensive Spatial Reasoning with Multimodal Large Language Model in Open Space},
      author={Weichen Zhang and Zile Zhou and Xin Zeng and Xuchen Liu and Jianjie Fang and Chen Gao and Yong Li and Jinqiang Cui and Xinlei Chen and Xiao-Ping Zhang},
      year={2025},
      eprint={2503.11094},
      archivePrefix={arXiv},
      primaryClass={cs.CV},
      url={https://arxiv.org/abs/2503.11094},
}

\clearpage
\appendix
\onecolumn

\section{Public Benchmark}\label{app:Public_Benchmark}
\begin{table}[H]
\centering
\scriptsize
\setlength{\tabcolsep}{2pt}
\renewcommand{\arraystretch}{1.3}
\begin{tabularx}{\textwidth}{|c|>{\hsize=1.16\hsize\raggedright\arraybackslash}X|>{\hsize=.84\hsize\raggedright\arraybackslash}X|l|c|l|c|c|}
\hline
\textbf{Modality} & \textbf{Task} & \textbf{Edge Application} & \textbf{Dataset} & \textbf{Device} & \textbf{Reference Model} & \textbf{Result (Q/L/E)} & \textbf{Metric} \\
\hline

\multirow{18}{*}{CV}
& \multirow{5}{*}{\mbox{Object detection\hspace{1.2pt}\cvtag{T01}\hspace{0.4pt}\cvtag{T02}\hspace{0.4pt}\cvtag{T03}\hspace{0.4pt}\cvtag{T04}\hspace{0.4pt}\cvtag{T05}}}
& Crop detection
& GlobalWheat~\cite{david2020globalWheat}
& NX & YOLO11n & 0.63/16.16/130.92 & mAP \\
\cline{3-8}
&  & Smart home monitoring
& HomeObjects-3K~\cite{Jocher_Ultralytics_Datasets_2025_HomeObjects_3K}
& NX & YOLOv8n FT & 0.41/250.15/-- & mAP \\
\cline{3-8}
&  & Autonomous driving
& KITTI~\cite{Geiger2013IJRR_KITTI}
& Orin & YOLOv8n & 0.63/8.89/-- & mAP \\
\cline{3-8}
&  & Aerial inspection
& DOTA~\cite{xia2018dota}
& Orin & YOLO11n-OBB & 0.28/10.83/0.11 & mAP \\
\cline{3-8}
&  & Common-object perception
& PASCAL VOC~\cite{everingham2010pascal_voc}
& NX & YOLOv8n & 0.79/15.71/-- & mAP@0.5 \\
\cline{2-8}

& \multirow{4}{*}{\mbox{Image classification\hspace{1.2pt}\cvtag{T06}\hspace{0.4pt}\cvtag{T07}\hspace{0.4pt}\cvtag{T08}\hspace{0.4pt}\cvtag{T09}}}
& \multirow{2}{*}{Edge visual analytics}
& CIFAR10~\cite{Krizhevsky09learningmultiple_cifar10_cifar100}
& TX2 & ResNet-20 & 0.92/6.20/-- & Accuracy \\
\cline{4-8}
&  &
& Caltech-101~\cite{fei2007learning_Caltech_101}
& TX2 & MobileNetV3 Linear & 0.90/42.89/-- & Accuracy \\
\cline{3-8}
&  & Steel-surface inspection
& NEU-CLS~\cite{song2013neu}
& TX2 & ResNet-18 Linear & 0.98/110.00/-- & Accuracy \\
\cline{3-8}
&  & Intelligent transportation
& GTSRB~\cite{Stallkamp-IJCNN-2011-gtsrb}
& TX2 & EfficientNet-B0 & 0.98/41.91/-- & Accuracy \\
\cline{2-8}

& Semantic segmentation \cvtag{T10}
& Road condition monitoring
& Crack Segmentation~\cite{crack-bphdr_dataset}
& NX & YOLO11n-seg & 0.80/709.84/-- & mIoU \\
\cline{2-8}

& \multirow{2}{*}{\mbox{Pose estimation\hspace{1.2pt}\cvtag{T11}\hspace{0.4pt}\cvtag{T12}}}
& Animal monitoring
& Dog-Pose~\cite{ultralytics_dog_pose,khosla2011fgvc}
& NX & YOLOv8n-Pose & 0.75/19.06/-- & PCK \\
\cline{3-8}
&  & Hand pose understanding
& Hand Keypoints~\cite{ultralytics_hand_keypoints}
& NX & YOLO11n-Pose & 0.90/18.96/0.15 & PCK \\
\cline{2-8}

& Video analytics \cvtag{T13}
& Drone-based analysis
& VisDrone~\cite{VisDrone}
& Orin & YOLOv8n & 0.11/8.97/0.10 & mAP \\
\cline{2-8}

& Face recognition \cvtag{T14}
& Face attribute analysis
& Adience~\cite{Eidinger2014AgeAG}
& TX2 & ResNet-18 & 0.84/16.06/-- & Accuracy \\
\cline{2-8}

& Optical character recognition \cvtag{T15}
& Document scanner
& Rendered SST2~\cite{radford2021learningtransferablevisualmodels}
& TX2 & ResNet-18 FT & 0.51/108.79/-- & Accuracy \\
\cline{2-8}

& Vehicle re-identification \cvtag{T16}
& Intelligent transportation
& VeRi-776~\cite{7553002}
& Orin & FastReID R50 & 0.82/127.03/-- & mAP \\
\cline{2-8}

& Crowd counting \cvtag{T17}
& Smart city surveillance
& ShanghaiTech~\cite{zhang2016shanghaitech}
& Orin & MobileNetV3-S Count & 119.80/2.88/-- & MAE \\
\cline{2-8}

& Anomaly detection \cvtag{T18}
& Industrial IoT (IIoT)
& MVTec AD~\cite{MVTecAD1,MVTecAD2}
& NX & MobileNetV3-S Features & 0.73/12.73/-- & AUROC \\
\hline

\multirow{15}{*}{NLP}
& \multirow{2}{*}{\mbox{Industrial AIOps\hspace{1.2pt}\nlptag{T19}\hspace{0.4pt}\nlptag{T20}}}
& Supercomputer analysis
& Loghub-BGL-2K~\cite{loghub1,loghub2}
& PC & Word-hash LR & 0.98/0.01/-- & F1 \\
\cline{3-8}
&  & Storage-system analysis
& Loghub-HDFS~\cite{loghub1,loghub2}
& PC & Hash LR & 0.91/0.01/-- & F1 \\
\cline{2-8}

& \multirow{2}{*}{\mbox{Intent classification\hspace{1.2pt}\nlptag{T21}\hspace{0.4pt}\nlptag{T22}}}
& Private assistant routing
& CLINC150~\cite{clinc150}
& PC & Word-bigram LR & 0.75/0.03/-- & Accuracy \\
\cline{3-8}
&  & Mobile banking support
& BANKING77~\cite{casanueva2020banking77}
& Pi5 & Word-hash LR & 0.89/0.05/-- & Accuracy \\
\cline{2-8}

& \multirow{2}{*}{\mbox{Text classification\hspace{1.2pt}\nlptag{T23}\hspace{0.4pt}\nlptag{T24}}}
& News filtering
& AG News~\cite{zhang2016ag-news}
& NX & DistilBERT & 0.94/506.89/-- & Accuracy \\
\cline{3-8}
&  & Product/service routing
& Ecommerce Text~\cite{gautam_2019_3355823}
& PC & Char-3gram LR & 0.94/0.40/-- & Accuracy \\
\cline{2-8}

& \multirow{3}{*}{\mbox{Text regression\hspace{1.2pt}\nlptag{T25}\hspace{0.4pt}\nlptag{T26}\hspace{0.4pt}\nlptag{T27}}}
& Medical review analysis
& WebMD Reviews~\cite{harode_webmd_drug_reviews_dataset}
& PC & Word-hash Ridge & 1.29/0.09/-- & RMSE \\
\cline{3-8}
&  & Writing feedback
& Feedback Prize ELL~\cite{feedback_prize_ell_2022}
& PC & BERT Mini & 1.00/14.40/-- & MCRMSE \\
\cline{3-8}
&  & Airline review scoring
& Airline Reviews~\cite{bhojani_airline_reviews}
& PC & Word-Bigram Ridge & 2.16/0.08/-- & RMSE \\
\cline{2-8}

& Semantic similarity \nlptag{T28}
& Local text matching
& GLUE STS-B~\cite{wang2019glue}
& PC & BERT Mini & 0.84/6.65/-- & Spearman \\
\cline{2-8}

& Spam classification \nlptag{T29}
& On-device SMS filtering
& SMS Spam Collection~\cite{sms_spam_collection_228}
& Pi5 & Word-2gram LR & 0.90/0.06/-- & F1 \\
\cline{2-8}

& \multirow{2}{*}{\mbox{Sentiment classification\hspace{1.2pt}\nlptag{T30}\hspace{0.4pt}\nlptag{T31}}}
& Social-text analysis
& TweetEval~\cite{barbieri2020tweeteval}
& Pi5 & BERT Mini & 0.66/6.70/-- & Accuracy \\
\cline{3-8}
&  & Review analysis
& SST-2~\cite{socher-etal-2013-recursive}
& Pi5 & BERT Mini & 0.85/9.34/-- & Accuracy \\
\cline{2-8}

& Emotion recognition \nlptag{T32}
& Conversational analytics
& GoEmotions~\cite{demszky2020goemotionsdatasetfinegrainedemotions}
& TX2 & RoBERTa & 0.47/1095.98/-- & Accuracy \\
\cline{2-8}

& Boolean QA \nlptag{T33}
& Lightweight QA
& BoolQ~\cite{clark2019boolq}
& NX & BERT Mini FT & 0.67/24.25/-- & Accuracy \\
\hline

\multirow{6}{*}{Audio}
& Speech recognition \audiotag{T34}
& Personal assistant
& LibriSpeech-100h~\cite{panayotov2015librispeech}
& Orin & Whisper & 0.06/223.21/-- & WER \\
\cline{2-8}
& \multirow{2}{*}{\mbox{Speech understanding\hspace{1.2pt}\audiotag{T35}\hspace{0.4pt}\audiotag{T36}}}
& Smart assistant
& FSC~\cite{lugosch2019fsc,benazir2024fsc}
& TX2 & Log-mel CNN & 0.35/4.45/18.26 & Accuracy \\
\cline{3-8}
&  & Personal assistant
& SLURP~\cite{SLURP}
& NX & HuBERT-SLU & 0.70/1668.41/-- & Accuracy \\
\cline{2-8}
& Emotion recognition \audiotag{T37}
& Local emotion analysis
& RAVDESS~\cite{livingstone2018ravdess}
& NX & Log-mel DS-CNN & 0.39/3.85/-- & Accuracy \\
\cline{2-8}
& Audio classification \audiotag{T38}
& Ambient sound analysis
& ESC-50~\cite{piczak2015dataset}
& TX2 & Log-mel ResNet-18 & 0.67/902.93/-- & Accuracy \\
\cline{2-8}
& Keyword spotting \audiotag{T39}
& Personal assistant
& Speech Commands~\cite{speechcommandsv2}
& Pi5 & Log-mel DS-CNN & 0.94/0.64/4.91 & Accuracy \\
\hline

\multirow{5}{*}{\shortstack{Sensing /\\Time-Series}}
& \multirow{2}{*}{\mbox{Human activity recognition\hspace{1.2pt}\sensetag{T40}\hspace{0.4pt}\sensetag{T41}}}
& Wearable sensing
& HHAR~\cite{hhar}
& Pi5 & DS-CNN & 0.80/0.08/0.53 & Accuracy \\
\cline{3-8}
&  & Smartphone sensing
& MotionSense~\cite{motionsense}
& Pi5 & Temporal 1D CNN & 0.94/0.06/0.44 & Accuracy \\
\cline{2-8}
& \multirow{3}{*}{\mbox{Time-series classification\hspace{1.2pt}\sensetag{T42}\hspace{0.4pt}\sensetag{T43}\hspace{0.4pt}\sensetag{T44}}}
& Industrial sensing
& EthanolConcentration~\cite{large2018ethanol}
& Pi5 & Spectral MLP & 0.27/0.35/-- & Accuracy \\
\cline{3-8}
&  & Physiological sensing
& Self-Regulation-SCP1~\cite{birbaumer1999spelling}
& Pi5 & ResNet-1D & 0.65/3.42/26.12 & Accuracy \\
\cline{3-8}
&  & Heartbeat monitoring
& Heartbeat~\cite{liu2016heart_sound_database}
& Pi5 & InceptionTime & 0.67/1.74/12.66 & Accuracy \\
\hline

\multirow{4}{*}{Tabular}
& \multirow{2}{*}{\mbox{Tabular classification\hspace{1.2pt}\tabtag{T45}\hspace{0.4pt}\tabtag{T46}}}
& Food quality grading
& Banana Quality~\cite{mrmars1010_banana_quality_dataset}
& PC & XGBoost & 0.75/0.08/-- & F1 \\
\cline{3-8}
&  & Software quality analysis
& Software Defects~\cite{cevik_software_defect_prediction}
& PC & Extra Trees & 0.73/0.01/-- & AUROC \\
\cline{2-8}
& \multirow{2}{*}{\mbox{Tabular regression\hspace{1.2pt}\tabtag{T47}\hspace{0.4pt}\tabtag{T48}}}
& Marine age estimation
& UCI Abalone~\cite{abalone_1}
& PC & Random Forest & 1.58/0.01/-- & MAE \\
\cline{3-8}
&  & Yield estimation
& Wild Blueberry Yield~\cite{qu2020wildblueberry}
& PC & HistGBR & 194.62/0.01/-- & MAE \\
\hline

\multirow{2}{*}{Multimodal}
& Audio captioning \mmtag{T49}
& Hearing accessibility
& Clotho~\cite{drossos2019clothoaudiocaptioningdataset,clotho_v2_1}
& Orin & EffNet-B2 Trm & 0.14/184.67/2190.03 & BLEU \\
\cline{2-8}
& Image captioning \mmtag{T50}
& Visual assistance
& Flickr8k~\cite{hodosh2013framing}
& Orin & BLIP & 0.19/3653.86/48170.06 & BLEU \\
\hline

\end{tabularx}
\vspace{5pt}
\caption{A comprehensive benchmark of edge-oriented on-device AI tasks. For each frozen Reference, the table reports task quality (Q), p95 latency in ms (L), and energy per inference in mJ (E). Energy constrains the search only for the 12 tasks with an application-level energy SLO; ``--'' denotes tasks without one.}
\label{tab:mcaas-benchmarks}
\end{table}

\clearpage
\twocolumn

\section{Benchmark and Evaluation Details}
\label{app:benchmark-details}
\enlargethispage{3\baselineskip}

\noindent\textbf{Reference selection and SLO construction.}
Because the selected baselines do not uniformly support this heterogeneous 50-task benchmark, frozen References provide common anchors for comparing task quality and device performance and constructing consistent Reference-relative SLOs.
Our Reference models are task-matched, task-adapted, and representative baselines drawn from classic model families and mature task-specific practices. We select and optimize our reference models and SLOs as follows.

\noindent $\bullet$ \textbf{Reference choice.} The 50-task benchmark contains 12 classical linear or tree models, 32 compact learned models, and six specialized pretrained models. The last group comprises FastReID, RoBERTa, Whisper, HuBERT, an EfficientNet-B2 captioner, and BLIP. This composition spans
lightweight edge models and task-specific pretrained models.
We draw our reference models from prominent open-source repositories and widely recognized platforms, including PapersWithCode~\cite{paperswithcode}, Kaggle~\cite{kaggle}, and HuggingFace~\cite{wolf2020huggingfacestransformersstateoftheartnatural}.

\noindent $\bullet$ \textbf{Reference optimization.} We train or fine-tune each model on the task’s training data, export its deployable artifact, and verify its predictions and runtime behavior on the assigned device. The References span modalities and optimized deployment stacks. For example, \inlinenlptag{T30} uses a fully fine-tuned BERT Mini exported to ONNX and measured across three Raspberry Pi~5 sessions. \inlinecvtag{T01} uses a trained YOLO11n compiled through a hardware-optimized TensorRT FP16 path on Xavier NX, achieving 16.16~ms p95 latency; \inlinesensetag{T40} uses a compact DS-CNN that achieves 0.08~ms and 0.53~mJ per inference on Raspberry Pi~5. These References therefore combine mature task recipes with deployment optimizations.

\noindent $\bullet$ \textbf{SLO definition.} For a controlled stress test, we define every SLO through the same relative ratio and freeze it before synthesis begins. Latency and active-energy limits are set to $0.8\times$ the corresponding Reference measurements, requiring \name{} to improve task quality while operating within physical budgets that are 20\% tighter than those of a strong deployable baseline.
The $0.8\times$ values are deliberate experimental stress targets: they normalize difficulty rather than represent a universal application deadline. This uniform rule removes per-task discretion; Table~\ref{tab:mcaas-benchmarks} retains the absolute quality, latency, and energy values, and Figure~\ref{fig:sensitivity-analysis} evaluates Reference-relative SLOs from $0.1\times$ to $1.0\times$.

Reference quality and deployment efficiency impose complementary challenges: a higher-quality Reference raises the bar for quality improvement, while faster and more energy-efficient deployment produces tighter absolute SLOs. Across classical, compact learned, and specialized pretrained References, \name{} achieves quality wins on 11/12, 25/32, and 4/6 tasks, and finds SLO-feasible artifacts on 10/12, 29/32, and 6/6 tasks, respectively. Thus, the results span the full Reference portfolio rather than concentrating in one model family.

\noindent\textbf{Latency and energy measurement.}
All latency and energy metrics are derived from repeated inference after warm-up. Latency is the p95 over repeated runtime invocations. Energy is measured independently of candidate-generated code and averaged over hardware-power samples aligned with the same repeated inference window. Jetson platforms use the on-board INA3221 sensor to sample VDD\_IN board-input power, while Raspberry Pi~5 uses \texttt{vcgencmd pmic\_read\_adc} to aggregate internal PMIC-rail power; both are sampled at 10~Hz.
This low sampling frequency does not conflict with sub-millisecond inference times. Each workload is sustained over a prolonged repeated-inference window, allowing the 10~Hz sampling to reliably capture steady-state active power.
Sampling begins before inference, records an idle window, and is accepted with at least three timestamp-aligned active samples. We compute energy per inference as mean active power multiplied by the timed duration and divided by the actual repetition count.
We derive measurement variation from the aligned samples, reporting total energy as the conservative primary metric and retaining idle-subtracted dynamic energy as an auxiliary metric.

\noindent\textbf{Dataset splits.}
We use the official dataset splits for all applicable methods.
Before each experiment, we freeze the training, validation, and test partitions. Model training uses the training partition, whereas our tree-based search uses only feedback from the validation partition. When an official test partition exists, it is excluded from the synthesis loop. We report the final performance on the held-out test partition.

\noindent\textbf{SEN evaluation details.}
We run three independent syntheses for each sample-level and cross-subject split, all using the same Raspberry Pi~5 SLOs, GPT-5.4 backend, 24-candidate budget, and verification pipeline. Validation results select one artifact per run, which is then evaluated on the held-out test partition. All six artifacts satisfy both physical SLOs: mean p95 latency is 0.251/0.352~ms and mean energy is 1.052/1.581~mJ for the sample-level/cross-subject settings. Table~\ref{tab:sen-generalizability} reports the mean and sample standard deviation across the three runs.

\noindent\textbf{MetaGPT-Verification.} This baseline preserves MetaGPT Data Interpreter’s
native planning, coding, debugging, and revision workflow. After each candidate,
an execution adapter trains it on the server, exports its declared artifact,
runs it on the target device, and returns task quality, runtime status, p95
latency, and applicable energy to the next revision. It uses the same GPT-5.4
backend, device and SLO contract, verifier, and per-task cap of 24 executable
candidates as \name{}. Thus, MetaGPT receives the same target-device evidence
while retaining its native linear search; tree control and cross-request rule
reuse remain the variables introduced by \name{}.

\clearpage
\balance
\section{Multi-Fidelity Verification and Failure Reuse}
\label{app:verifier-reproducibility}

\noindent\textbf{Multi-Fidelity eligibility and cold start.}
The fast measurement (P1) may reject a candidate only when an earlier full
verification covers the same model graph, device, runtime, precision, input
profile, and measurement setup. The first candidate under a new setting
therefore proceeds to full verification; subsequent candidates under that
setting may use its calibrated fast measurement. A change in any matching field
returns the candidate to full verification. The safety evaluation spans
Reference-relative SLOs from $0.01\times$ to $0.8\times$. All 60 early
rejections agree with full verification, corresponding to a one-sided exact
95\% upper bound of 4.87\% on the disagreement rate. The calibration margin
also includes the observed measurement variation.

\noindent\textbf{Verified failure rules and matching.}
A rule represents a reproduced device–runtime incompatibility rather than an
error string alone. It combines the failure stage and normalized error with the
device, runtime version, artifact format, and the implicated condition, such as
an IR version, opset, data type, operator, shape, or precision setting. A rule
becomes active only after the same failure is reproduced on its target device and
runtime, and it skips a candidate only when these fields match. The 50 rules
cover TX2, Xavier NX, Orin AGX, and Raspberry Pi~5 (23/6/9/12 rules), as well as
TensorRT, ONNX Runtime, and LiteRT (16/32/2). They represent operator, data-type,
dynamic-output, IR-version, schema, and build/load failures
(10/4/2/29/3/2). Twenty-five rules also match an artifact from a different
dataset, showing that reuse is not limited to one task. Once established,
deterministic artifact-version limits are enforced as P0 checks before device
execution. Operator-, data-type-, and dynamic-output conditions remain verified
failure rules because they depend on the artifact structure and its
device--runtime context.

\begin{table}[H]
\centering
\footnotesize
\caption{Evaluation scope and aggregate outcomes for the two
verification mechanisms.}
\label{tab:verifier-summary}
\setlength{\tabcolsep}{3pt}
\renewcommand{\arraystretch}{1.05}
\begin{tabular}{@{}p{0.24\columnwidth}p{0.70\columnwidth}@{}}
\toprule
\textbf{Measure} & \textbf{Evaluation and outcome} \\
\midrule
\multicolumn{2}{@{}l}{\textbf{Multi-Fidelity verification}} \\
Verification cost & 16 fixed candidates across four runtimes: per-runtime
time falls by 17.6–39.7\%, and total time falls from 69.13 to 51.35 minutes
(25.7\%), with the same selected artifact. \\
Applicability & Of 180 candidates, 120 yield matched fast/full measurements;
69 have prior calibration at decision time. \\
Decision safety & 60 early rejections across the evaluated SLO settings;
full verification agrees with all 60. \\
\addlinespace[2pt]
\multicolumn{2}{@{}l}{\textbf{Verified Failure Reuse}} \\
Coverage & 134 candidates from 15 datasets, five modalities, four devices, and three runtimes; all 50 rules are reused, matching 57 candidates. \\
Device cost & Required device evaluations fall from 101 to 44 (56.4\%). \\
Non-IR benefit & After deterministic IR-version checks at P0, the remaining
rules reduce required device evaluations from 69 to 44 (36.2\%). \\
Decision safety & All 57 matched candidates reproduce the predicted failure under device execution. \\
\bottomrule
\end{tabular}
\end{table}

\noindent\textbf{Rule-bank warm-up and workload shifts.}
The service can warm the rule bank on representative workloads and continue to
add rules as devices and runtimes evolve. A workload, device, or runtime shift
reduces how often existing rules match until the bank covers the new setting;
it does not broaden the scope of an existing rule. Candidates outside that
scope proceed to device execution, where a reproduced incompatibility can
establish a rule for the new setting.

\end{document}